%% file: main_neurips2022.tex
\documentclass{article}

\PassOptionsToPackage{round}{natbib}

\usepackage[final]{neurips_2022}

\usepackage[utf8]{inputenc} %
\usepackage[T1]{fontenc}    %
\usepackage{hyperref}       %
\usepackage{url}            %
\usepackage{booktabs}       %
\usepackage{amsfonts}       %
\usepackage{nicefrac}       %
\usepackage{microtype}      %
\usepackage{xcolor}         %

\usepackage{microtype}
\usepackage{graphicx}
\usepackage{subfigure}
\usepackage{booktabs} %
\usepackage{lipsum}		%
\usepackage{graphicx}
\usepackage{natbib}
\usepackage{doi}
\usepackage{authblk}

\usepackage{multibib}
\newcites{appendix}{Appendix References}

\input{custom/packages.tex}
\input{custom/abbrevs.tex}
\usepackage{hyperref}

\title{Invariance Learning in Deep Neural Networks
with Differentiable Laplace Approximations}

\author[1,2]{Alexander Immer\thanks{Equal contribution, order decided by coin flip.
Correspondence to: \texttt{alexander.immer@inf.ethz.ch}, \texttt{tycho.vanderouderaa@imperial.ac.uk}}$^{*,}$}
\author[3]{Tycho F.A. van der Ouderaa$^{*,}$}
\author[1]{\authorcr Gunnar R\"atsch}
\author[1,4]{Vincent Fortuin}
\author[3]{Mark van der Wilk}
\affil[1]{Department of Computer Science, ETH Zurich, Switzerland}
\affil[2]{Max Planck Institute for Intelligent Systems, T\"ubingen, Germany}
\affil[3]{Department of Computing, Imperial College London, UK}
\affil[4]{Department of Engineering, University of Cambridge, UK}

\everypar={\looseness=-1}
\let\svthefootnote\thefootnote
\newcommand\blfootnote[1]{%
  \let\thefootnote\relax%
  \footnotetext{#1}%
  \let\thefootnote\svthefootnote%
}

\begin{document}

\maketitle

\begin{abstract}
\input{sections/abstract.tex}
\end{abstract}

\section{Introduction}
\label{sec:introduction}
\input{sections/introduction.tex}
\section{Related Work}
\label{sec:related-work}

\input{sections/related_work.tex}
\section{Background}
\label{sec:background}
\input{sections/background.tex}
\section{Invariance Learning with Differentiable Laplace Approximations}
\label{sec:method}
\input{sections/method.tex}

\section{Experiments}%
\label{sec:experiments}
\input{sections/experiments.tex}

\section{Discussion and Limitations}%
\label{sec:discussion}
\input{sections/discussion.tex}

\section{Conclusion}%
\label{sec:conclusion}
\input{sections/conclusion.tex}
\section*{Acknowledgements}
A.I.\ acknowledges funding by the Max Planck ETH Center for Learning Systems (CLS).
V.F.\ acknowledges funding by the Swiss Data Science Center through a PhD Fellowship, the Swiss National Science Foundation through a Postdoc.Mobility Fellowship, St John's College Cambridge through a Research Fellowship, and the Branco Weiss Foundation through a Branco Weiss Fellowship.

\bibliography{bibliography.bib}  %
\bibliographystyle{plainnat}

\newpage

\clearpage
\appendix
\onecolumn

\section{Comparison with Related Approaches}
\label{app:method_comparison}

\begin{table*}[h]
\caption{Overview of related approaches and our proposed method. Note that only Augerino~\citep{benton2020learning} and our method work for deep neural networks, while using only training data in a single run.
Our method has the additional benefit of being Bayesian, parameterisation-independent, and requires no additional hyperparameter, which empirically can lead to better performance. We discuss the corresponding issues of Augerino in detail in \cref{app:augerion-comparison}}
\label{table:overview_methods}
\begin{center}
\resizebox{\linewidth}{!}{
\begin{tabular}{lccccc}\toprule
Approach & \makecell{single\\run} & \makecell{train data\\only} & \makecell{deep neural\\network} & \makecell{Bayesian\\justification} & \makecell{parameterisation-\\independent}
\\ \midrule
\citet{cubuk2018autoaugment} & \xmark & \xmark & \cmark & \xmark & \cmark \\
\citet{zhou2020meta} & \xmark & \xmark & \cmark & \xmark & \cmark \\
\citet{lorraine2020} & \xmark & \xmark & \cmark & \xmark & \cmark \\
\citet{van2018learning} & \cmark & \cmark & \xmark & \cmark & \cmark \\
\citet{schwoebel2022layer} & \cmark & \cmark & \xmark & \cmark & \cmark \\
\midrule
\citet{benton2020learning} & \cmark & \cmark & \cmark & \xmark & \xmark \\
\textbf{this work} & \cmark & \cmark & \cmark & \cmark & \cmark \\
\bottomrule
\end{tabular}
}
\end{center}
\end{table*}

\section{Affine Invariance Parameterisation}
\label{app:technical-appendix}
\input{sections/appendices/technical_details}

\section{Failure cases of Augerino}
\label{app:augerion-comparison}
\input{sections/appendices/augerino_comparison}

\section{Computational Complexities}
\label{app:complexity}
\input{sections/appendices/complexity}

\section{Details on Efficient Gradient Estimation with \ggn and \kfac}
\label{app:jvp}
\input{sections/appendices/jvp}
\section{Discussion of Approximations}
\label{app:approximations}
\input{sections/appendices/approximations}

\section{Mechanism of Differentiable Laplace for Invariance Learning}
\label{app:mechanism_intuition}
\input{sections/appendices/mechanism}

\section{Detailed Algorithm}
\label{app:algorithm}
\input{sections/appendices/algorithm}

\newpage
\section{Training Details}
\label{app:training-details}
\input{sections/appendices/training_details}

\section{Additional Results and Experiments}
\label{app:experiments}

\subsection{Classification Example}
\label{app:toy_example}
\input{sections/appendices/toy_examples}

\subsection{Quantitative Results}
\label{app:quantitative-results}
\input{sections/appendices/quantitative_results}

\newpage
\subsection{Additional Results: Invariance Learning Trajectories on All Datasets}
\label{app:additional-augmentation-trajectories}
\input{sections/appendices/additional_trajectories}
\newpage
\subsection{Additional Results: Learned Invariance Barplots and Samples on all Datasets}
\label{app:additional-learned-invariances}
\input{sections/appendices/additional_barplots}

\newpage
\subsection{Additional Results: Subset Experiments on All Datasets}
\label{app:additional-subset-experiments}
\input{sections/appendices/additional_subsets}

\end{document}

%% file: custom/packages.tex
\usepackage{microtype}
\usepackage{graphicx}
\usepackage{float}
\usepackage{color}
\usepackage{url}
\usepackage{rotating}
\usepackage{adjustbox}
\usepackage{wrapfig}
\usepackage{algorithm}
\usepackage{algpseudocode}

\usepackage{etoolbox}
\usepackage{booktabs}
\usepackage{multirow}
\usepackage{multicol}
\usepackage{array}
\usepackage{enumitem}
\newcolumntype{C}{>{$}c<{$}}
\newcolumntype{L}{>{$}l<{$}}
\newcolumntype{R}{>{$}r<{$}}
\setitemize[0]{leftmargin=*,itemsep=0.0pt,topsep=0.0pt,partopsep=0.0pt,parsep=0.0pt}

\usepackage{amsmath}
\usepackage{amsfonts}
\usepackage{bbm}
\usepackage{mathtools}
\usepackage{xspace}
\usepackage{nicefrac}
\usepackage{makecell}

\usepackage[capitalise,nameinlink]{cleveref}
\crefname{section}{Sec.}{Secs.}
\crefname{appendix}{App.}{Apps.}
\crefname{algorithm}{Alg.}{Algs.}
\creflabelformat{equation}{#2\textup{#1}#3}  %

\usepackage{tikz}
\usetikzlibrary{positioning}
\usetikzlibrary{calc}
\usetikzlibrary{arrows.meta}
\usetikzlibrary{backgrounds}
\usetikzlibrary{decorations.pathmorphing, patterns}
\usetikzlibrary{decorations.pathreplacing}
\usetikzlibrary{plotmarks}

\usepackage{contour}

\definecolor{lightgray}{gray}{0.85}
\definecolor{lightlightgray}{gray}{0.9}
\definecolor{C1}{HTML}{1F77B4}
\definecolor{C2}{HTML}{FF7F0E}
\definecolor{C3}{HTML}{2CA02C}
\definecolor{C4}{HTML}{D62728}
\definecolor{C5}{HTML}{9467BD}
\colorlet{C1light}{C1!70!white}
\colorlet{C2light}{C2!70!white}
\colorlet{C3light}{C3!70!white}
\colorlet{C4light}{C4!70!white}
\colorlet{C5light}{C5!70!white}
\colorlet{C1lighter}{C1!40!white}
\colorlet{C2lighter}{C2!40!white}
\colorlet{C3lighter}{C3!40!white}
\colorlet{C4lighter}{C4!40!white}
\colorlet{C5lighter}{C5!40!white}
\colorlet{C1vlight}{C1!20!white}
\colorlet{C2vlight}{C2!20!white}
\colorlet{C3vlight}{C3!20!white}
\colorlet{C4vlight}{C4!20!white}
\colorlet{C5vlight}{C5!20!white}

\usepackage[normalem]{ulem}

\usepackage{pifont}
\newcommand{\cmark}{\textcolor{C3}{\ding{51}}}
\newcommand{\xmark}{\textcolor{C4}{\ding{55}}}

%% file: custom/abbrevs.tex
\newcommand{\param}{\ensuremath{\vtheta}\xspace}

\newcommand{\paramstar}{{\vtheta_\ast}}
\newcommand{\model}{\ensuremath{\sM}\xspace}

\newcommand{\augmentation}{\ensuremath{\boldsymbol{\eta}}\xspace}
\newcommand{\augmentationstar}{\ensuremath{\boldsymbol{\eta_\ast}}}

\newcommand{\hypothesis}{\sH}

\newcommand{\vxaug}{\ensuremath{\vx'\xspace}}
\newcommand{\vfaug}{\ensuremath{\hat{\vf}\xspace}}
\newcommand{\vxn}{\vx_n}
\newcommand{\vyn}{\vy_n}

\newcommand{\diffaug}{\tfrac{\partial}{\partial \eta}}
\newcommand{\ourmethod}{\textsc{lila}\xspace}

\newcommand{\ggn}{\textsc{ggn}\xspace}

\newcommand{\kfac}{\textsc{kfac}\xspace}
\newcommand{\ef}{\textsc{ef}\xspace}

\newcommand{\sminus}{\text{-}\xspace}

\newcommand{\jac}{\ensuremath{\mJ_{\vtheta}(\vx)}\xspace}
\newcommand{\jacn}{\ensuremath{\mJ_{\vtheta}(\vx_n)}\xspace}

\newcommand{\network}{\ensuremath{\vf(\vx, \vtheta)}\xspace}

\newcommand{\pms}[1]{\ensuremath{{\scriptstyle\pm #1}}}
\newcommand{\wrt}{w.r.t.\xspace}

\newcommand{\transpose}{^\mathrm{\textsf{\tiny T}}}

\newcommand{\kron}{\otimes}

\DeclareMathOperator{\mkvec}{vec}

\newcommand{\inv}{^{-1}}
\newcommand{\defeq}[0]{\mathrel{\stackrel{\textnormal{\tiny def}}{=}}}
\newcommand{\E}{\mathbb{E}}

\newcommand{\R}{\mathbb{R}}

\def\sB{{\mathcal{B}}}

\def\sD{{\mathcal{D}}}

\def\sH{{\mathcal{H}}}

\def\sM{{\mathcal{M}}}

\def\D{{\sD}}
\def\data{{\sD}}

\def\vzero{{\mathbf{0}}}

\def\va{{\mathbf{a}}}

\def\vf{{\mathbf{f}}}
\def\vg{{\mathbf{g}}}
\def\vh{{\mathbf{h}}}

\def\vv{{\mathbf{v}}}

\def\vx{{\mathbf{x}}}
\def\vy{{\mathbf{y}}}

\def\vtheta{{\boldsymbol{\theta}\xspace}}

\def\vlambda{{\boldsymbol{\lambda}\xspace}}

\def\vepsilon{{\boldsymbol{\epsilon}\xspace}}

\def\mA{{\mathbf{A}}}

\def\mG{{\mathbf{G}}}
\def\mH{{\mathbf{H}}}
\def\mI{{\mathbf{I}}}
\def\mJ{{\mathbf{J}}}

\def\mLambda{{\boldsymbol{\Lambda}\xspace}}

\newcommand{\half}{\mbox{$\frac{1}{2}$}}

\newcommand{\given}{\mbox{$|$}}

\newcommand{\order}[1]{\ensuremath{\mathcal{O}(#1)}}
\newcommand{\simiid}{\overset{{\scriptscriptstyle\mathrm{iid}}}{\sim}}

\newcommand{\rnd}[1]{\left(#1\right)}
\newcommand{\sqr}[1]{\left[#1\right]}

\newcommand{\tikzunderlinewithtext}[3]{%
    \contourlength{0.08em}
    \tikz[baseline=(tounderline.base)]{
        \node[inner sep=0pt,outer sep=0pt] (tounderline) {\contour{white}{#2}};
        \useasboundingbox ($(tounderline.base west)+(0, 0)$) rectangle (tounderline.base east);
        \begin{scope}[on background layer]
            \draw[#1] ($(tounderline.base west)-(0,0.1)$) -- ($(tounderline.base east)-(0,0.1)$) node[pos=0.5, below] {\textcolor{black}{#3}};
        \end{scope}
    }%
}%
\newcommand{\mathunderlinewithtext}[2]{
    \tikzunderlinewithtext{decorate,decoration={brace,mirror}, line width=0.5pt, black}{$#1$}{{\scriptsize $#2$}}}

%% file: sections/abstract.tex
Data augmentation is commonly applied to improve performance of deep learning by enforcing the knowledge that certain transformations on the input preserve the output. Currently, the data augmentation parameters are chosen by human effort and costly cross-validation, which makes it cumbersome to apply to new datasets. We develop a convenient gradient-based method for selecting the data augmentation without validation data during training of a deep neural network. Our approach relies on phrasing data augmentation as an invariance in the prior distribution on the functions of a neural network, which allows us to learn it using Bayesian model selection. This has been shown to work in Gaussian processes, but not yet for deep neural networks. We propose a differentiable Kronecker-factored Laplace approximation to the marginal likelihood as our objective, which can be optimised without human supervision or validation data. We show that our method can successfully recover invariances present in the data, and that this improves generalisation and data efficiency on image datasets.\blfootnote{The code is available at \url{https://github.com/tychovdo/lila}}

%% file: sections/introduction.tex
Data augmentation is a commonly used machine learning technique that is essential to high-performing deep learning and computer vision systems. 
It aims to obtain a model that is \textit{invariant} to a set or distribution of transformations, by fitting a model with inputs that are transformed in a way that is known to leave the output class unchanged.
This procedure can be regarded as artificially creating more data and is well known to increase generalisation performance and data efficiency.
Yet, choosing the right transformations is an expensive and task-specific process that relies on domain knowledge and human effort, as well as trial-and-error through cross-validation. 
This can quickly become intractable when many parameters are considered, particularly if they are continuous, because each setting requires training a model to convergence.

We aim to make selecting suitable transformations easier, by learning them via gradient descent. %
Our approach is inspired by the procedure of \citet{van2018learning}, which casts learning invariances and data augmentations as a Bayesian model selection problem. 
This view suggests selecting invariances by maximising the \emph{marginal likelihood} with gradient-based optimisation. While this approach was successful in Gaussian process models, it has not yet been demonstrated in deep neural networks, where the marginal likelihood is harder to approximate.

\begin{figure}[t]
  \centering
  \includegraphics{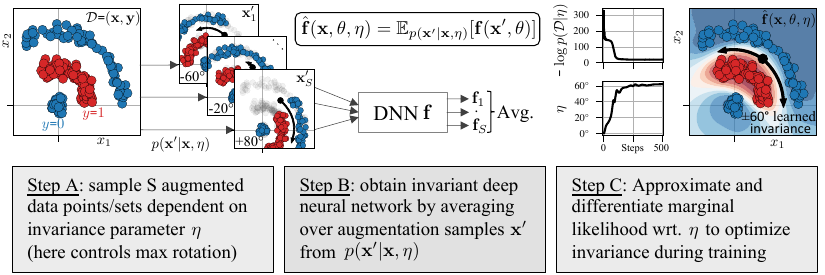}
  \vskip -0.5em
    \caption{Illustration of our approach on a dataset with rotational invariance.
    Steps~A~and~B allow to define a Bayesian neural network with a likelihood dependant on invariance parameter $\eta$~\citep{van2018learning}.
    Our contributions enable tractable marginal likelihood estimation and differentiation during training, as illustrated in Step~C.
    The right-most figure shows the posterior predictive of our Bayesian neural network after learning the invariance present in the data.}
  \vskip -1em
  \label{fig:illustration}
\end{figure}

To circumvent this problem, we built upon the scalable Laplace approximation to the marginal likelihood developed by \citet{immer2021scalable}, who recently showed that maximising it can successfully select neural network hyperparameters, such as architectures. 
We extend their method to enable gradient-based optimisation of complex hyperparameters that control invariances in deep neural networks. 
To that end, we propose an efficient and differentiable Kronecker-factored Laplace approximation for invariant neural networks and a novel method to obtain stochastic gradients with respect to invariance parameters, which is also useful for optimising other hyperparameters. %
Our method is the first to enable differentiable Bayesian model selection to learn complex hyperparameters, in particular invariances, in deep neural networks.

Our approach is illustrated in \cref{fig:illustration}.
We specify invariances as a parameterised distribution over perturbations of the network's input, like in data augmentation~(Step A). 
The model output is averaged over samples from the distribution~(Step B), which yields a Bayesian neural network with likelihood dependent on the perturbations~\citep{nabarro2021data}. 
We derive a marginal likelihood approximation for such neural networks and show how to efficiently compute its gradients with respect to the invariance parameters~(Step C). By approximating the marginal likelihood, we can differentiably learn invariances during training, jointly with neural network parameters, and without validation data.

We demonstrate experimentally that our method can differentiably learn useful distributions over affine invariances, which are common data augmentations, on various versions of the image classification datasets MNIST, FashionMNIST, and CIFAR-10, without validation data. 
Our learned invariances improve the generalisation and data efficiency of neural networks, without the effort required for choosing data augmentations or custom architectures.
On original datasets, our method can increase test performance by up to $8\sminus 10$ percentage points. On random subsets of image classification datasets, we show that our method can achieve up to $10\times$ better data efficiency. 
Our work strengthens how Bayesian methods can be useful for deep learning beyond predictive uncertainty estimation.

%% file: sections/related_work.tex
\paragraph{Invariances in deep learning.}
Since the inception of the convolutional neural network (CNN) \citep{fukushima1982neocognitron,lecun1998gradient}, building invariances and equivariances into deep learning models has drastically increased data efficiency and generalisation~\citep{cohen2018spherical, brandstetter2021geometric}, for instance, on image classification~\citep{cohen2016group}, molecular dynamics~\citep{batzner2021se}, and reinforcement learning~\citep{van2020mdp}.
However, these approaches require knowing the invariances a priori.
In this work, on the other hand, we aim to automatically learn the correct type and amount of invariance from data without supervision.

\vspace{-0.8em}
\paragraph{Learning invariances.}
Learning invariances from data is hard, because symmetries define constraints on the functions a network can represent, and therefore do not improve data fit according to the training loss, even if they would lead to better generalisation on test data. 
Some methods have therefore proposed to learn invariances or data augmentation by estimating gradients on the validation loss. 
For example, AutoAugment~\citep{cubuk2018autoaugment} learns data augmentation using policy gradients,
\citet{lorraine2020} use the implicit function theorem,
and \citet{zhou2020meta} phrase the problem as meta-learning.
Such approaches require validation sets of sufficient size to prevent high variance and overfitting \citep{lorraine2020}.
Here, we tackle the problem of learning invariances when validation data is not available.

\vspace{-0.8em}
\paragraph{Validation-free invariance learning.}
When casting invariance learning as a Bayesian model selection problem, we can select helpful invariances using the marginal likelihood of the model with training data alone.
This has been successfully demonstrated in Gaussian processes \citep{van2018learning} and in the weight space of single-layer neural networks \citep{van2021learning}. To scale this approach to deep networks, \citet{schwoebel2022layer} attempt to use a marginal likelihood that is only computed in the last layer.
While the latter was successful for small neural networks, it failed to learn invariances on more complex datasets that require deep neural networks,
likely due to known limitations of last-layer approaches \citep{ober2021promises,amersfoort2021}.
Augerino~\citep{benton2020learning} is an alternative to Bayesian model selection that works for deep learning by regularising invariances to increase during training.
However, this approach depends on the chosen parameterisation of the invariance and may still require a validation set to tune the regularisation strength.
In \cref{app:augerion-comparison}, we show that these issues can be detrimental to Augerino's performance. 
In contrast, our proposed method uses the Bayesian marginal likelihood estimate in deep neural networks that is parameterisation-invariant and improves performance.
The main differences between our proposed method and the alternatives are summarised in~\cref{app:method_comparison}.

\vspace{-0.6em}
\paragraph{Bayesian model selection for deep learning.} \looseness=-1
Marginal likelihood approximations have recently enabled gradient-based hyperparameter optimisation for deep neural networks~\citep{immer2021scalable, ober2021global, antoran2022probabilistic}.
Laplace approximations~\citep{mackay1992practical} with structured Hessian approximations, such as \kfac~\citep{martens2015optimizing, botev2017practical}, or linearisation have been shown to improve generalisation, for example, by optimising weight regularisation~\citep{immer2021scalable, daxberger2021laplace} and learning length-scales of convolutions~\citep{antoran2022probabilistic}. 
Other approximate inference methods that rely on ensembling or sampling do not directly apply to marginal likelihood estimation while common methods like mean-field variational inference~\citep{blundell2015weight} or last-layer approaches~\citep{ober2021promises, daxberger2021laplace, schwoebel2022layer} have been shown to fail for model selection.
We therefore focus on Laplace approximations and extend them to enable gradient-based optimisation of more complex hyperparameters like invariances.

%% file: sections/background.tex
\vspace{-0.2em}
We consider a supervised learning task with dataset $\D = \{(\vx_n, \vy_n)\}_{n=1}^N$ consisting of $N$ inputs $\vx \in \R^D$ and targets $\vy \in \R^C$.
Our goal is to learn the function $\vf: \R^D \to \R^C$ that relates the inputs and outputs, which we represent as a neural network $\vf(\vx;\param)$ with parameters $\param$. We control which solutions for $\vf$ are preferred over others (i.e.,~the \emph{inductive bias}) with hyperparameters $\hypothesis$, which parameterise the prior over $\param$ and $\vf$.
Invariance is a particularly helpful inductive bias that constrains the output of $\vf$ to remain similar for certain transformations of the input $\vx$,
which can improve generalisation by allowing a single datapoint to inform predictions for a wider range of inputs.
Our goal is to learn useful invariances together with the network weights.

\subsection{Parameterising Invariance}
To construct and parameterise invariant functions, we consider a local invariance that intuitively requires the function to not change ``too much'' in response to transformed inputs.
To obtain such an invariant function $\vfaug$, we average an unconstrained function $\vf$ over a perturbation distribution $p(\vx' \given \vx, \augmentation)$.
In practice, we sample $\vxaug$ by reparameterising $\vepsilon \sim p(\vepsilon)$ with a differentiable function $\vg$ and approximate the expectation with $S$ Monte Carlo samples $\vepsilon_{1},\ldots,\vepsilon_{S} \simiid p(\vepsilon)$~\citep{van2018learning, benton2020learning}:
\begin{align}
    \vfaug(\vx; \param, \augmentation) = \E_{p(\vx'|\vx,\augmentation)} [\vf(\vx'; \param)]
    = \E_{p(\vepsilon)} [ \vf(\vg(\vx,\vepsilon;\augmentation);\param)]
    \approx \tfrac{1}{S} {\textstyle \sum_{s}} \vf(\vg(\vx,\vepsilon_s;\augmentation);\param)
    \,,
    \label{eq:invariant_function}
\end{align}
where $\vfaug$ is differentiable in the parameters $\augmentation$ that control the perturbation distribution and therefore the invariance.
When the perturbation distribution is uniform on the orbit of a group, we recover exact invariance in $\vfaug$ \citep{kondorthesis,ginsbourger2016}.
The perturbation distribution resembles data augmentation applied to $\vf$ instead of the loss and is also used at test time ($\vfaug$ depends on it).
Because the unconstrained function $\vf$ is a neural network, we refer to $\vfaug$ as an \textbf{invariant neural network}.

Given that we can parameterise a suitable data augmentation distribution $p(\vxaug \given \vx, \augmentation)$, we can learn invariances in arbitrary domains or group structures.
What parameterisation works best in practice remains a research question. 
In our experiments, we consider a combination of uniform distributions and corresponding parameters $\augmentation \in \R^6$  over $6$ generator matrices that define a probability density over the group of affine transformations, similar to \citet{benton2020learning} and detailed in \cref{app:technical-appendix}.

Finding invariance parameters $\augmentation$ is hard because $\vfaug$ is a constrained version of $\vf$ and, especially for flexible models, this cannot improve the data fit according to the standard training loss~(see also \cref{app:maximum-likelihood}). 
To overcome this, we propose to use Bayesian inference which provides a convenient framework to optimise invariance parameters with gradients during training without validation data.

\subsection{Bayesian Model Selection}
Bayesian inference prescribes how unknowns, such as invariance parameters, should be determined from data. 
To infer hyperparameters $\hypothesis$ from data, we are interested in the their posterior, $p(\hypothesis \given \data)\propto p(\data \given \hypothesis) p(\hypothesis)$~\citep{mackay2003information}.
Because this posterior is intractable, type II maximum likelihood (ML-II) is often used instead, which obtains a point estimate $\hypothesis^\ast$ for the optimal hyperparameters according to the \emph{marginal likelihood} $p(\data \given \hypothesis)$, which requires integration over the model parameters.

ML-II is routinely used in Gaussian processes \citep[\S~5.2]{rasmussen2006gaussian} and trades off model simplicity with data fit \citep{rasmussen2001occam}. 
There are also strong relations to quantities from other statistical frameworks, like cross-validation \citep{fong2020marginal}, Minimum Description Length \citep{grunwald2007minimum}, and generalisation error bounds \citep{germain2016pac}.
\Citet{van2018learning} showed its usefulness for selecting invariances, and elaborated on the mechanism by which it works. The main advantage of the marginal likelihood, 
\begin{align}
  \label{eq:marglik}
  p(\data|\hypothesis) 
  = \textstyle\int p(\data\given \param, \hypothesis) p(\param \given \hypothesis) \, \mathrm{d}\param 
  = \textstyle\int \prod_{n=1}^N p(\vy_n|\vf(\vx_n;\param),\hypothesis) p(\param|\hypothesis) \, \mathrm{d} \param \,, 
\end{align}
is that it can be computed from the training data alone and optimised with gradients. 
In our work, the integration is over the neural network parameters and requires particularly scalable approximations.

\subsection{Bayesian Model Selection for Deep Learning}
Computing the marginal likelihood for neural networks involves intractable integrals. 
The Laplace approximation \citep[\S~27]{laplace1774memoire,mackay2003information} offers a solution by approximating the log joint of the parameters and data, $p(\data, \param \given \hypothesis)$, with a second-order Taylor expansion around a mode $\paramstar$:
\begin{align}
  \log p(\data \given \hypothesis) 
  \approx \log p(\data, \paramstar \given \hypothesis) - \half \log \left\vert\tfrac{1}{2\pi} \mH_{\paramstar} \right\vert ,  \label{eq:lap_marglik}
\end{align}
where the first term decomposes into the log likelihood, a sum over data points, and the log prior, both of which are cheap and easy to evaluate and correspond to a typical training loss evaluated at a mode $\paramstar$.
The second term depends on the log determinant of the log-joint Hessian at the same mode
\vspace{-0.5em}
\begin{equation}
    \mH_{\paramstar} \defeq -\nabla_{\param}^2 \log p(\data, \param \given \hypothesis)\vert_{\param=\paramstar} 
    = \mH^{\textsc{nll}}_{\paramstar} -\nabla_{\param}^2 \log p(\param \given \hypothesis)\vert_{\param=\paramstar} , 
\end{equation}
where $\mH^{\textsc{nll}}_{\paramstar}$ denotes the Hessian of the negative log likelihood.
This approach allows to estimate the marginal likelihood using a MAP estimate $\paramstar$ of the weights and its local curvature $\mH_{\paramstar}$. 

To circumvent the high cost of estimating the full Hessian,
structured generalised Gauss-Newton (\ggn) approximations are preferred for model selection in deep learning, as also in optimisation~\citep{martens2014new, bottou2010large}.
\citet{immer2021scalable} recently demonstrated successful hyperparameter and architecture selection with such approximations.
Further, they observe empirically that their algorithm does not require to be at a mode $\paramstar$, which allows for interleaved gradient-based optimisation of parameters and hyperparameters during training.
With the Jacobian matrix 
$\jac \defeq \tfrac{\partial \vf(\vx;\param)}{\partial \param} \in \R^{C \times P}$ 
of the network output given input $\vx$ w.r.t.~parameters, and Hessian of the log likelihood w.r.t.~network outputs $\mLambda(\vf) = - \nabla_{\vf}^2 \log p(\vy \given \vf)$,
the \ggn simplifies the negative-log-likelihood-dependent term of the Hessian:
\vspace{-0.5em}
\begin{align}
  \mH_\param^{\textsc{nll}}
  \approx 
  \mH^\ggn_\param 
  \defeq \, &{\textstyle\sum_{n=1}^N} \jacn\transpose \mLambda(\vf(\vx_n;\param)) \jacn. \label{eq:ggn}
\end{align}
Here, we assume that $\vf$ forms the natural parameters of an exponential family likelihood function~\citep{murphy2012machine}.
In classification, for example, $\vf$ are the logits.
We refer to the resulting approximations as Laplace-\ggn. 
To overcome the still intractable quadratic size of $\mH^\ggn_\param$ in $P$, \citet{immer2021scalable} use structured approximations like \kfac~\citep{martens2015optimizing}. 
Cheaper approximations, such as diagonal ones, often compromise accuracy~\citep[\cref{app:approximations};][]{daxberger2021laplace}. %

\subsection{Kronecker-Factored Gauss-Newton Approximation (\kfac)}

Kronecker-factored approximations to the Gauss-Newton, such as \kfac, are commonly used for Laplace approximations as they currently seem to provide the best known trade-off between performance and complexity~\citep{ritter2018scalable, daxberger2021laplace}.
\kfac is a block-diagonal approximation to the Gauss-Newton matrix $\mH^\ggn_\param$ where each block corresponds to a layer in the neural network~\citep{martens2015optimizing, botev2017practical}. 
It is particularly efficient because each block is represented as two Kronecker factors instead of one dense matrix.
For example, the \ggn block of a layer with $D \times G$ parameters, that is, a fully-connected layer connecting $D$ to $G$ neurons, would have quadratic memory complexity \order{D^2 G^2} while the corresponding Kronecker approximation of \kfac is in \order{D^2 + G^2} and is therefore even tractable for wide neural networks.

Mathematically, \kfac approximates the \ggn of the $l$th block of the neural network parameters by enforcing a Kronecker factorisation across data points. %
This would only be exact for a single data point $\vx_n$, for which we can write the \ggn block corresponding to the parameters of the $l$th layer as
\begin{align}
    \mH_{l,n}^\ggn 
    = 
    [\va_{l,n} \kron \vg_{l,n}]\mLambda_n[\va_{l,n} \kron \vg_{l,n}]\transpose 
   =
   [\va_{l ,n} \va_{l, n}\transpose] \kron [\vg_{l,n}\mLambda_n \vg_{l,n}\transpose] 
   \defeq \mA_{l, n} \kron \mG_{l, n},
   \label{eq:kfac_ggn_start} 
\end{align}
where $\va_{l,n} \in \R^{D_l}$ is the input to the $l$th layer for data point $\vx_n$, $\vg_{l,n} \in \R^{G_l \times C}$ is the transposed Jacobian of the network output with respect to the output of the $l$th layer for $\vx_n$, and $\mLambda_n = \mLambda(\vf(\vx_n;\param))$.
Thus, the factors in \cref{eq:kfac_ggn_start} are the Jacobian terms as in the \ggn (\cref{eq:ggn}) but only for the $l$th layer, i.e.,~$\mJ_{\param_l}(\vx_n)\transpose = \va_{l,n} \kron \vg_{l,n}$.
\kfac then approximates the sum over $N$ data points by summing up the Kronecker factors individually instead of breaking the Kronecker-factored structure:
\begin{align}
    \mH_{l}^\ggn &= {\textstyle \sum_{n=1}^N} [\va_{l ,n} \va_{l, n}\transpose] \kron [\vg_{l,n}\mLambda_n \vg_{l,n}\transpose] 
    \approx \tfrac{1}{N} \sqr{\mathunderlinewithtext{{\textstyle \sum_{n=1}^N} \va_{l ,n} \va_{l, n}\transpose}{\defeq \mA_{l}}} \kron \sqr{\mathunderlinewithtext{{\textstyle \sum_{n=1}^N} \vg_{l,n}\mLambda_n \vg_{l,n}\transpose}{\defeq \mG_l}}, \label{eq:kfac_general}
\end{align}

where $\mA_{l} \in \R^{D_l \times D_l}$ and $\mG_l \in \R^{G_l \times G_l}$ can be understood as the uncentered covariance over $N$ data points of the inputs to the $l$th layer and the Jacobians wrt.\ the output of the $l$th layer, respectively.
The normalization by $\tfrac{1}{N}$ is necessary to account for the additional terms that arise from distributing the sum over the factors.
For a single-layer model, i.e.,~a linear model, \kfac is exact~(cf.~\cref{app:approximations}).

%% file: sections/method.tex
We propose a Laplace-\ggn approximation to the marginal likelihood for invariant neural networks and enable gradient-based optimisation of their invariance parameters during training, without the use of validation data~(see \cref{fig:illustration} for a high-level overview).
In our approach, we integrate the augmentation distribution $p(\vxaug\given \vx, \augmentation)$ into a Bayesian neural network model such that the marginal likelihood directly depends on the invariance parameters $\augmentation \in \hypothesis$.
In particular, this is due to a modified likelihood function $p(\vy \given \vfaug(\vx; \param, \augmentation), \hypothesis)$~\citep[c.f.,~][]{nabarro2021data}.
This model enables optimisation of the invariance parameters~\augmentation using gradient ascent on the log marginal likelihood,
\begin{equation}
  \augmentation \gets \augmentation + \nabla_{\augmentation} \log p(\data \given \hypothesis = \{\augmentation, \model\}) \,, \label{eq:invariance_gradient_ascent}
\end{equation}
where $\model=\hypothesis\setminus \{\augmentation\}$ are remaining hyperparameters, such as regularisation strength or model architecture.
However, the tractable Laplace-\ggn and \kfac approximations are not available for invariant Bayesian neural networks, which prohibits straightforward application of the update in~\cref{eq:invariance_gradient_ascent} using the methods described in~\cref{sec:background}.

In the following, we extend the Laplace-\ggn and \kfac approximations to invariant neural networks~(\cref{sec:marglik_invariant,sec:kfac_aug}), which enables optimising the log marginal likelihood in parallel to the neural network parameters, as in \citet{immer2021scalable}.
However, their algorithm has an intractable memory complexity for computing the gradients w.r.t.\ invariance parameters or other complex hyperparameters that act on the neural network function $\vf$ directly.
In practice, they only considered gradient-based optimisation of hyperparameters that act linearly on the Hessian $\mH_\param$, for example regularisation strength and observation noise. 
In \cref{sec:grad_aug}, we lift this constraint by proposing a method to obtain gradients w.r.t.\ complex hyperparameters without memory overhead.
The final algorithm and a discussion of approximations for invariance learning are detailed in \cref{app:algorithm,app:approximations}.

\subsection{Laplace-\ggn for an Invariant Neural Network}\label{sec:marglik_invariant}
To define the Laplace-\ggn approximation to the log marginal likelihood, the hyperparameter objective, we need to extend the \ggn to invariant neural networks.
For an invariant neural network with parameter estimate $\paramstar$, the log marginal likelihood approximation is given by
\begin{align}
  \log p(\D \given \augmentation, \model)
  \approx\ &
  \textstyle{\sum_{n=1}^N} \log p(\vyn \given \vfaug(\vxn;\paramstar,\augmentation), \model)
  + \log p(\paramstar \given \model)
  - \tfrac{1}{2} \log \vert \tfrac{1}{2\pi} \hat{\mH}_{\paramstar}^\ggn(\augmentation) \vert, \label{eq:aug_laplace_ggn}
\end{align}
where the first two terms constitute the training loss of the invariant neural network corresponding to the log joint as in the vanilla Laplace approximation in \cref{eq:lap_marglik} and 
the last term is the Gauss-Newton approximation $\hat{\mH}^\ggn_{\paramstar}$ of the invariant neural network $\vfaug$, which we derive below.
The first and last term depend on the invariance parameter $\augmentation$ that we want to learn and differentiate with respect to.
We use $S$ Monte Carlo samples $\vepsilon_{1},\ldots,\vepsilon_{S} \simiid p(\vepsilon)$ to estimate the invariant neural network $\vfaug$ as in \cref{eq:invariant_function}.

\textbf{The log-likelihood} term is approximated with $S$ Monte Carlo samples, which leads to a lower bound,
\begin{align}
  \textstyle{\sum_{n=1}^N} \log p(\vyn \given \vfaug(\vxn;\param,\augmentation), \model) \nonumber
  &\geq
  \textstyle{\sum_{n=1}^N} \E_{\vepsilon_{1},\ldots,\vepsilon_{S}} [\log p(\vyn \given \tfrac{1}{S} \textstyle{\sum_s} \vf(\vg(\vx_n,\vepsilon_s;\augmentation); \param), \model)]
  \\
  &\approx
  \textstyle{\sum_{n=1}^N} \log p(\vyn \given \tfrac{1}{S} \textstyle{\sum_s} \vf(\vg(\vx_n,\vepsilon_s;\augmentation); \param), \model)\,,
  \label{eq:aug_model_sampled}
\end{align}
which is due the concavity of the log likelihood in its natural parameter $\vfaug$ and Jensen's inequality as shown by \citet{nabarro2021data}, \citet{schwoebel2022layer}, and detailed in \cref{app:approx_list}.
The subsequent Monte Carlo approximation is then unbiased. Increasing $S$ leads to a tighter bound and improves the approximation. 
In practice, we sample $\vepsilon_s$ independently per data point $\vxn$ to reduce correlation.
We can obtain a stochastic gradient w.r.t.~\augmentation by sampling a mini-batch of $M \ll N$ data but it is not possible to batch over the $S$ augmentations that parameterise the likelihood function.
The runtime and memory complexity are therefore increased by a factor of~$S$.
For a subset of data and $S \leq 100$ augmentations, the gradient \wrt $\augmentation$ can be computed using backpropagation~\citep{benton2020learning}.

\textbf{The Gauss-Newton} can be derived from the log-likelihood approximation in \cref{eq:aug_model_sampled} using the same $S$ samples.
In particular, the Jacobian and log-likelihood Hessian required for the \ggn are given by
\begin{equation}
    \hat{\mJ}_\param(\vx;\augmentation) \defeq \tfrac{1}{S} {\textstyle \sum_{s}}  \mJ_\param(\vg(\vx,\vepsilon_s;\augmentation))
    \;\;\textrm{and}\;\;
    \hat{\mLambda}(\vx;\param,\augmentation) \defeq \mLambda(\tfrac{1}{S} {\textstyle \sum_{s}} \vf(\vg(\vx,\vepsilon_s;\augmentation);\param)).
\end{equation}
Both terms depend on the invariance parameter \augmentation and are later differentiated with respect to it.
The resulting \ggn of the invariant neural network log likelihood estimated with $S$ samples is defined as
\begin{align}
\label{eq:aug_ggn}
  \hat{\mH}^\ggn_\param (\augmentation)
  &\defeq 
  {\textstyle\sum_{n=1}^N \hat{\mJ}_\param (\vxn;\augmentation)\transpose \, \hat{\mLambda}(\vx_n;\param,\augmentation) \, \hat{\mJ}_\param (\vxn;\augmentation) }.
\end{align}
The \ggn for an invariant model therefore requires averaging individual Jacobians and functions of the underlying neural network $\vf$. 
This is in contrast to the \ggn of an improper Bayesian model~\citep{wenzel2020good} with standard data augmentation, which averages log-likelihood terms instead of functions.
In this case, entire \ggn terms are averaged over the $S$ augmentations and the marginal likelihood cannot be optimised because it requires tempering~\citep{immer2021scalable}.
Computing the \ggn of an invariant model increases the runtime by a factor of~$S$ over a non-invariant model.
The empirical Fisher, which is often cheaper, can be extended analogously and requires averaging gradients instead of Jacobians.
Finally, the Laplace-\ggn is obtained by computing the log determinant.
The Laplace-\ggn approximation derived here already allows to learn small invariant neural networks with few layers and small datasets via automatic differentiation.
For example, this is tractable for the classification example illustrated in \cref{fig:illustration}.
However, \emph{computing} the log determinant of the \ggn has a cubic complexity in the number of parameters $P$ and is therefore intractable for deep neural networks on larger datasets. To enable its estimation, we extend \kfac to invariant neural networks in \cref{sec:kfac_aug}.
Further, \emph{differentiating} the log determinant has intractable memory complexity because it does not allow for a stochastic gradient but requires construction of a computational graph for the entire \ggn approximation.
This is a key limitation of the method proposed by \citet{immer2021scalable} when optimising complex hyperparameters, for which we provide a solution in \cref{sec:grad_aug}.

\subsection{Extending \kfac to Invariant Neural Networks}
\label{sec:kfac_aug}

Computing \kfac as described in \cref{sec:background} for an invariant neural network would not preserve the Kronecker structure and therefore be intractable for deep neural networks due to the quadratic cost in the numbers of parameters per layer.
In particular in \cref{eq:kfac_ggn_start}, \kfac uses the fact that the Jacobian w.r.t.\ the parameters of the $l$th layer, $\mJ_{\param_l}(\vx_n)$, can be written as the Kronecker product $\va_{l,n} \kron \vg_{l, n}$.
For an invariant neural network, the Kronecker structure cannot be maintained because of the sum over $S$ augmentation-sample Jacobians, each of which constitutes a Kronecker product: 
\begin{equation}
    \hat{\mJ}_{\param_l}(\vx_n; \augmentation) =
    \tfrac{1}{S} {\textstyle \sum_s} \mJ_{\param_l}(\vg(\vx_n,\vepsilon_s;\augmentation))
    = \tfrac{1}{S}{\textstyle \sum_s} \big[\va_{l,n}^{(s)} \kron \vg_{l,n}^{(s)}\big]
    \label{eq:inv_kfac_jac}
    \, ,
\end{equation}
where $\va^{(s)}, \vg^{(s)}$ depend on the $s$th sample $\vx_s = \vg(\vx,\vepsilon_s;\augmentation)$ and thus \augmentation.
In general, the sum of Kronecker products does not allow for efficient computation and requires evaluating the products, which is intractable as it requires \order{D_l^2G_l^2} instead of \order{D_l^2+G_l^2} memory.

Similar to the idea underlying \kfac itself~\citep{martens2015optimizing}, we enforce the efficient Kronecker-factored structure by approximating the sum of a Kronecker product as a Kronecker product of sums and appropriately normalizing by the number of terms.
However, instead of applying this approximation to the \ggn over multiple data points, we apply it to the Jacobian and have
\begin{equation}
    \hat{\mJ}_{\param_l}(\vx_n; \augmentation) 
    = \tfrac{1}{S}{\textstyle \sum_s} [\va_{l,n}^{(s)} \kron \vg_{l,n}^{(s)}]
    \approx \tfrac{1}{S^2} \big[{\textstyle \sum_s} \va_{l,n}^{(s)}\big] \kron \big[{\textstyle \sum_s} \vg_{l,n}^{(s)}\big]
    \label{eq:inv_jac_kfac}
    \, .
\end{equation}
Applying this Jacobian approximation, the $n$th summand of the \ggn can be written as
\begin{align}
    \hat{\mH}_{l,n}^\ggn (\augmentation)
    &\approx 
    \Big[\tfrac{1}{S^2} \big[{\textstyle \sum_s} \va_{l,n}^{(s)}\big] \kron \big[{\textstyle \sum_s} \vg_{l,n}^{(s)}\big] \Big]
    \hat{\mLambda}(\vx_n;\param,\augmentation)
    \Big[\tfrac{1}{S^2} \big[{\textstyle \sum_s} \va_{l,n}^{(s)}\big] \kron \big[{\textstyle \sum_s} \vg_{l,n}^{(s)}\big] \Big]\transpose
    \label{eq:kfac_aug}
    \\
    &= \textstyle{
    \tfrac{1}{S^4} \sqr{\big[\sum_s \va_{l,n}^{(s)}\big] \big[\sum_s \va_{l,n}^{(s)}\big]\transpose} \kron
    \sqr{\big[\sum_s \vg_{l,n}^{(s)}\big] 
    \hat{\mLambda}(\vx_n;\param,\augmentation)
    \big[\sum_s \vg_{l,n}^{(s)}\big]\transpose}
    }
    \defeq \hat{\mA}_{l,n} \kron \hat{\mG}_{l,n}.
    \nonumber
\end{align}
To compute the full \kfac, we then accumulate the Kronecker factors for the $l$th layer, $\hat{\mA}_{l,n}$ and $\hat{\mG}_{l,n}$, of the invariant neural network over all $n$ data points to obtain $\hat{\mA}_l$ and $\hat{\mG}_l$ as in \cref{eq:kfac_general} for vanilla \kfac. 
Like \kfac, our approximation for invariant models remains exact for linear models~(\cref{app:approx_list}).

The log determinant required for the marginal-likelihood approximation can be computed from $\hat{\mG}_l$, $\hat{\mA}_l$ individually~\citep{immer2021scalable} and has a complexity of \order{D_l^3+G_l^3} as opposed to the intractable \order{D_l^3G_l^3}.
This enables us to apply our method to deep invariant neural networks with widths of order $10^4$, like vanilla \kfac.
We discuss computational complexities in depth in \cref{app:complexity}.
We note that \kfac for invariant neural networks could further be of independent interest for second-order optimisation and inference~\citep{martens2015optimizing, zhang2018noisy}.

\subsection{Efficient Gradient Estimation of the Laplace-\ggn w.r.t.\ Complex Hyperparameters}\label{sec:grad_aug}
Automatic differentiation~(\textsc{ad}) of the log determinant term in the Laplace-\ggn w.r.t.\ complex hyperparameters, such as invariances, has an intractable memory complexity.
Here, we propose a method to estimate the gradient of the log determinant in the Laplace-\ggn without memory overhead.
For \textsc{ad}, the \emph{memory} complexity is equivalent to the \emph{runtime} complexity of computing the log determinant, which is at least \order{NP}, the cost of training a deep neural network for one epoch.
Such computation is only tractable due to batching (e.g.,~for standard training losses) and otherwise would require several terabytes of memory for deep neural networks.
However, the log determinant does not allow for a batched gradient and therefore \textsc{ad} requires storing the full training data pass.

Our approach to reducing the memory complexity relies on computing a vector-Jacobian product where both, the vector and the Jacobian, can be estimated from batches of data.
Mathematically, the problem reduces to differentiation of the log determinant of a sum of square matrices w.r.t. hyperparameter~$\augmentation$, i.e.,~$\tfrac{\partial}{\partial \augmentation} \log \vert \mH(\augmentation) \vert$ with $\mH(\augmentation) = \sum_n \mH_n(\augmentation) \in \R^{P \times P}$. 
For a positive definite matrix $\mH$, we have $\tfrac{\partial \log \vert \mH \vert}{\partial \mH}=\mH\inv$. %
Therefore, we can differentiate w.r.t.~$\augmentation$ with
\begin{align}
    \tfrac{\partial}{\partial \augmentation} \log \vert \mH(\augmentation) \vert 
    &= \textstyle \sum_{p=1}^P \sum_{q=1}^P [\mH\inv(\augmentation)]_{p,q} [\tfrac{\partial}{\partial \augmentation} \sum_{n=1}^N \mH_n(\augmentation)]_{p,q} \nonumber \\
    &= \mkvec\rnd{\mH\inv(\augmentation)}\transpose \textstyle \sum_{n=1}^N \tfrac{\partial}{\partial \augmentation} \mkvec \rnd{\mH_n(\augmentation)}\,,
    \label{eq:aug_grad_eff}
\end{align}
where the two vectorised matrices are $P^2$-dimensional vectors and can both be computed from individual batches of $\mH_n$ as follows:
the first term acts as a \emph{preconditioner} and can be computed by summing up the batches and inverting the resulting matrix $\mH$ without storing the computation graph.
The second term is a sum over $N$ Jacobians \wrt~\augmentation and we can either aggregate it or obtain an unbiased stochastic estimate from batches of data.
The product of both terms constitutes a vector-Jacobian product and is a standard procedure of \textsc{ad}~\citep{paszke2017automatic}.
\begin{figure*}[t]
    \centering
    \includegraphics[width=\textwidth]{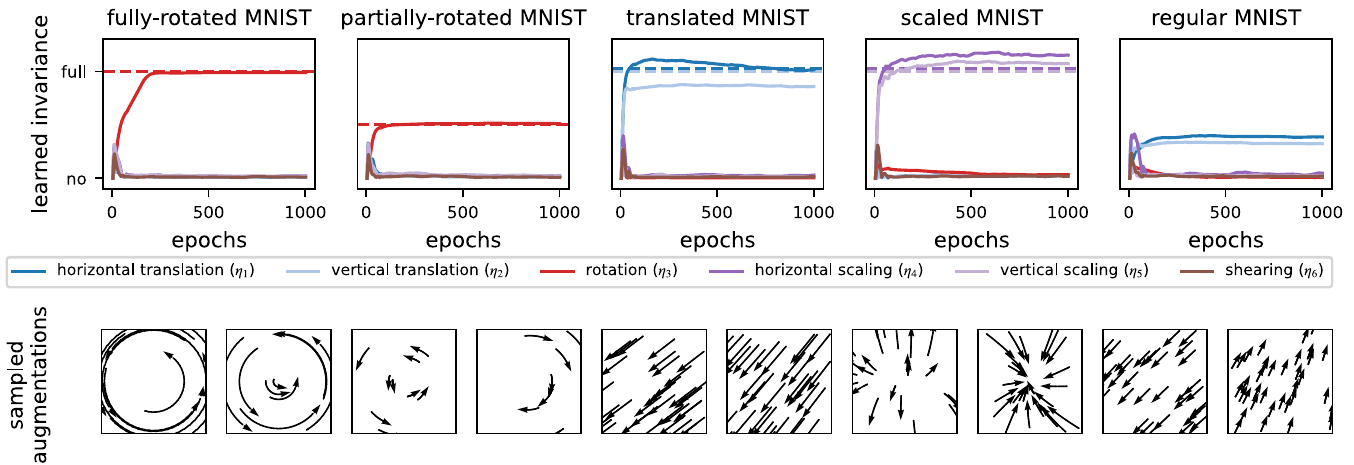}
    \vspace{-1.5em}
    \caption{Trajectories of invariance parameters $\boldsymbol{\eta}=[\eta_1, \eta_2, ..., \eta_6]^T$ optimised with \ourmethod over $1000$ epochs for different versions of the MNIST dataset. From each distribution, two sampled transformations are represented as flow fields with arrow lengths scaled to pixel movement. \ourmethod automatically learns the correct type and amount of invariance for each of the modified training datasets and keeps the other affine invariance parameters at zero as desired.} %
    \label{fig:invariance-trajectories}
\end{figure*}

The proposed method allows to aggregate gradients with respect to complex hyperparameters with memory complexity that is controlled by the batch size $1 \leq M \leq N$. 
In contrast to naive application of \textsc{ad} to the log determinant, this allows to decouple memory and runtime complexity and
enables gradient-based optimisation of the log determinant in \cref{eq:aug_laplace_ggn} w.r.t.\ the augmentation parameters for deep invariant networks on large datasets.
More generally, our method enables gradient-based marginal likelihood optimisation for more complex hyperparameters than previously considered~\citep{immer2021scalable, antoran2022probabilistic}.
In \cref{app:jvp}, we describe the gradient computation for \kfac in detail.

%% file: sections/experiments.tex
We evaluate our method that \emph{learns invariances using Laplace approximations} (\ourmethod) by optimising affine invariances on different MNIST~\citep{lecun2010mnist}, FashionMNIST~\citep{FashionMNIST}, and CIFAR-10~\citep{krizhevsky2009learning} classification tasks. 
To validate whether the method is capable of learning appropriate invariances, we construct several additional datasets modified by known sets of invariance transformations with the goal to recover them. 
We consider the following affine transformations: full rotation, partial rotation, translation, and scaling (full details in \cref{app:technical-appendix}). We compare our approach with a non-invariant baseline and Augerino \citep{benton2020learning}, which is, to our knowledge, the only other method that is capable of learning invariances on complex image datasets in deep neural networks without validation data. 
In the non-invariant model, prior parameters are learned using the marginal likelihood~\citep{immer2021scalable}. For invariance learning, prior parameters and $\boldsymbol{\eta}$ are jointly learned based on the marginal likelihood. For Augerino, we minimise the regular cross entropy with added regularising term $-10^{-2} ||\boldsymbol{\eta}||_2$ and a fixed weight decay of $10^{-4}$, following the original paper \citep{benton2020learning}. The same parameterisation, network architecture, and initialisations (in particular $\boldsymbol{\eta}$ $=$ $\boldsymbol{0}$) were used for all methods. We assess performance of our approach by inspecting learned invariances $\boldsymbol{\eta}$, marginal likelihoods, and final test performances.

\subsection{Recovering Known Invariances}
\label{sec:recovering-learned-invariances}

To assess the invariances learned by our method \ourmethod , we can inspect the learned invariance parameters. The invariance parameter vector $\boldsymbol{\eta} = [\eta_1, \eta_2, ..., \eta_6]^T$ describes affine invariances with components corresponding to x-translation, y-translation, rotation, horizontal and vertical scaling, and shearing~(\cref{app:technical-appendix}). As an MLP has little symmetry encoded in the architecture itself, we expect $\augmentation$ to almost correctly recover the invariances. In \cref{fig:invariance-trajectories}, we plot the trajectories of each vector component over the course of training for an MLP model on different transformed MNIST datasets. As reference, we show the amount of invariance that was imposed on each dataset as a dashed line. From the figure, we can observe that for each dataset, \ourmethod learns the correct invariance as well as the amount of each invariance. To some extent, the model also learned translational invariance on the regular MNIST dataset which can be explained by intrinsic translational invariance of the dataset. 

Other network architectures have certain symmetries already built-in to some extent (e.g., translational equivariance of convolutional layers). Yet, we find that \ourmethod is also capable of inferring the correct invariances with such larger and other network architectures and on a variety of datasets in \cref{app:additional-augmentation-trajectories}.
In \cref{app:approximations}, we further show that cheaper Hessian approximations, such as the diagonal \ggn instead of \kfac and empirical Fisher (\ef) instead of \ggn lead to worse performance for invariance learning.
This suggests that our extension of \kfac to invariant neural networks is necessary for sufficient invariance learning.
While \citet{immer2021scalable} find that diagonal approximations can suffice for learning regularisation hyperparameters, this does not apply for the more complex invariance parameters considered here and more accurate approximations tend to increase performance.

\begin{figure*}[t]
    \centering
    \includegraphics[width=\textwidth]{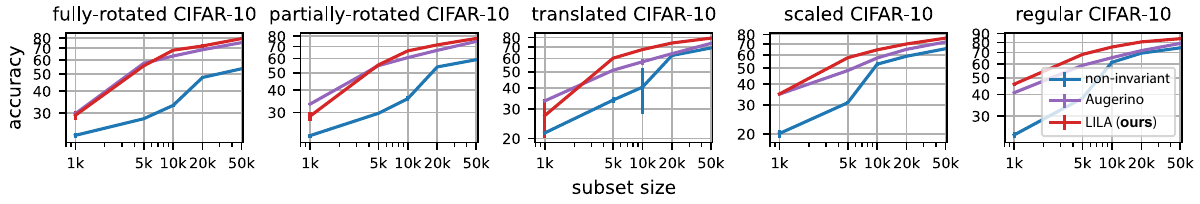}
    \vspace{-2em}
    \caption{\ourmethod improves data efficiency on different versions of CIFAR-10 by learning invariances. We compare test accuracy on randomly sampled subsets to a non-invariant network and Augerino using a ResNet. For example on fully-rotated CIFAR-10, invariance learning achieves the final accuracy of the non-invariant model with 10 times less data. In most cases, \ourmethod further improves over Augerino. We plot the average performance and standard error over three seeds.}
    \label{fig:subset-plot}
\end{figure*}
\input{sections/table_accuracy_only.tex}

\subsection{Invariance Learning in Different Networks}
\label{sec:different-models}
\begin{wraptable}{r}{0.4\textwidth}
    \centering
    \vspace{-3em}
    \begin{adjustbox}{max width=0.4\textwidth}
    \begin{tabular}{l|c }
      Method
      & Test accuracy
      \\
    \hline
    non-invariant
 & 85.17 \pms{0.39}
      \\
    Augerino
 & 87.67 \pms{0.08} 
      \\
    \ourmethod (\textbf{ours})
 & \textbf{91.98 \pms{0.04}}
      \\
    \ourmethod \textsc{ef} (\textbf{ours})
     & 91.64 \pms{0.26}\\
    \hline
    \end{tabular}
    \end{adjustbox}
    \caption{Invariance Learning with \kfac on CIFAR-10 with a Wide ResNet achieves best test accuracy. Also, the cheaper empirical Fisher (\ef) variant improves over Augerino.}
    \label{tab:large-resnet-results}
    \vspace{-0.5em}
\end{wraptable}
To quantify the benefit of models with learned invariances, we present final test accuracies of \ourmethod and the baselines with different models on each of the datasets in \cref{tab:quantitative-table-accuracy-only}.
Additional marginal likelihood scores can be found in \cref{app:quantitative-results}. 
In terms of test accuracy and marginal likelihood, 
we find that learning invariances always outperforms the non-invariant baseline and that our approach improves over Augerino in almost all cases. 
This holds across the transformed and original datasets. 
While the improvements are modest for MNIST, the performance improvements on F-MNIST and CIFAR-10 can be up to 10 percent points.
In \cref{tab:large-resnet-results}, we use the commonly used Wide ResNet architecture on CIFAR-10 and find that invariance learning with Augerino merely improves performance while our method achieves performance improvement of almost 7\% points. 
We also report the performance of \ourmethod with the cheaper \kfac-\ef instead of \ggn.

\subsection{Invariance Learning Improves Data Efficiency}
Invariances can be particularly useful for data-efficiency, which can be evaluated by measuring performance on subsets of data. 
In \cref{fig:subset-plot}, we show the test accuracy of \ourmethod, the non-invariant baseline, and Augerino trained on different subsets of CIFAR-10 and its modified versions. We further provide results for all architectures and datasets in \cref{app:additional-subset-experiments}.
In general, we find that invariance learning with both Augerino and our method always improves performance across datasets and subset sizes. Most notably on a subset of 1000 regular CIFAR-10 dataset samples, Augerino improved performance by 18 percentage points compared to the non-invariant baseline, whereas our method showed an improvement of 22 percentage points. \ourmethod requires only 10\% of the data to obtain the same accuracy as the non-invariant model on the fully-rotated dataset. These findings suggest that learning invariances is useful in general, and in particular when limited data is available.

\subsection{On the Learned Distributions}
In some instances (see CIFAR-10 results in \cref{app:additional-augmentation-trajectories,app:additional-learned-invariances}), we observed that the model learned translational invariance on the scaled dataset rather than scale invariance. Our independent uniform distributions over invariance components cannot capture correlations between components whereas the scaled dataset was jointly scaled across the horizontal and vertical axes and thus is correlated. We hypothesise that more complex distributions that do allow for capturing correlations could offer a potential solution in such cases, but leave investigation of more complex families to future work.

%% file: sections/table_accuracy_only.tex
\begin{table*}
    \centering
    \begin{adjustbox}{max width=1.0\linewidth}%
    \begin{tabular}{l l l|c c c c c}
      Dataset
      & Network
      & Method
      & {Fully Rotated}
      & {Partially Rotated}
      & {Translated}
      & {Scaled}
      & {Original}
      \\
    \hline
    \multirow{6}{*}{MNIST} & 
    \multirow{3}{*}{MLP} & 
    \multicolumn{1}{|l|}{non-invariant}
 & 93.82 \pms{0.10} & 95.83 \pms{0.03} & 94.15 \pms{0.02} & 97.07 \pms{0.06} & 98.20 \pms{0.03}
      \\
    &
    & 
    \multicolumn{1}{|l|}{Augerino}
 & \textbf{97.83 \pms{0.03}} & 96.35 \pms{0.02} & 94.47 \pms{0.08} & 97.45 \pms{0.03} & 98.45 \pms{0.03}
      \\
    &
    & 
    \multicolumn{1}{|l|}{Diff. Laplace (\textbf{ours})}
 & 97.74 \pms{0.07} & \textbf{97.81 \pms{0.11}} & \textbf{97.28 \pms{0.05}} & \textbf{98.33 \pms{0.05}} & \textbf{98.98 \pms{0.05}}
 \\
    \cline{2-8}
    &
    \multirow{3}{*}{CNN} & 
    \multicolumn{1}{|l|}{non-invariant}
 & 95.97 \pms{0.33} & 97.51 \pms{0.17} & 96.54 \pms{0.29} & 98.37 \pms{0.00} & 99.09 \pms{0.02}
      \\
    & 
    &
    \multicolumn{1}{|l|}{Augerino}
 & \textbf{99.04 \pms{0.02}} & 98.91 \pms{0.03} & 97.79 \pms{0.09} & 98.77 \pms{0.06} & 98.26 \pms{0.10}
      \\
    &
    & 
    \multicolumn{1}{|l|}{\textsc{lil} \kfac (\textbf{ours})}
 & 98.83 \pms{0.07} & \textbf{98.92 \pms{0.05}} & \textbf{98.69 \pms{0.07}} & \textbf{99.01 \pms{0.06}} & \textbf{99.42 \pms{0.02}}
      \\
    \hline
    \multirow{6}{*}{F-MNIST} & 
    \multirow{3}{*}{MLP} & 
    \multicolumn{1}{|l|}{non-invariant}
 & 77.62 \pms{0.30} & 81.10 \pms{0.23} & 77.68 \pms{0.10} & 81.84 \pms{0.05} & 88.48 \pms{0.56}
      \\
    &
    & 
    \multicolumn{1}{|l|}{Augerino}
 & 77.76 \pms{0.15} & 81.40 \pms{0.05} & 78.05 \pms{0.10} & 82.46 \pms{0.09} & 89.10 \pms{0.13}
      \\
    &
    & 
    \multicolumn{1}{|l|}{\ourmethod (\textbf{ours})}
 & \textbf{87.39 \pms{0.06}} & \textbf{86.72 \pms{0.13}} & \textbf{84.62 \pms{0.08}} & \textbf{84.31 \pms{0.06}} & \textbf{89.94 \pms{0.12}}
      \\
    \cline{2-8}
    &
    \multirow{3}{*}{CNN} & 
    \multicolumn{1}{|l|}{non-invariant}
 & 78.69 \pms{0.28} & 82.12 \pms{0.35} & 80.33 \pms{0.19} & 83.66 \pms{0.37} & 89.54 \pms{0.23}
      \\
    & 
    &
    \multicolumn{1}{|l|}{Augerino}
 & 85.76 \pms{3.23} & 81.54 \pms{0.19} & 82.94 \pms{0.13} & 83.58 \pms{0.08} & 90.07 \pms{0.12}
      \\
    &
    & 
    \multicolumn{1}{|l|}{\ourmethod (\textbf{ours})}
 & \textbf{89.45 \pms{0.03}} & \textbf{88.40 \pms{0.00}} & \textbf{87.73 \pms{0.20}} & \textbf{87.33 \pms{0.00}} & \textbf{91.92 \pms{0.21}}
      \\
    \hline
   \multirow{3}{*}{CIFAR-10} & 
    \multirow{3}{*}{ResNet} &  %
    \multicolumn{1}{|l|}{non-invariant}
 & 54.16 \pms{0.40} & 59.90 \pms{0.12} & 69.65 \pms{0.16} & 66.06 \pms{0.13} & 74.13 \pms{0.51}
      \\
    & 
    &
    \multicolumn{1}{|l|}{Augerino}
 & 75.40 \pms{0.19} & 74.76 \pms{0.34} & 73.71 \pms{0.31} & 72.07 \pms{0.09} & 79.03 \pms{1.04}
      \\
    &
    & 
    \multicolumn{1}{|l|}{\ourmethod (\textbf{ours})}
 & \textbf{79.50 \pms{0.62}} & \textbf{77.71 \pms{0.46}} & \textbf{79.21 \pms{0.17}} & \textbf{76.03 \pms{0.15}} & \textbf{84.19 \pms{0.76}}
      \\
    \hline
    \end{tabular}
    \end{adjustbox}
    \caption{Test accuracy for models using \ourmethod on different versions of the MNIST, FashionMNIST, and CIFAR-10 datasets. We report the average performance and the standard error over three random seeds. Our method outperforms the non-invariant network and Augerino for most models and datasets.}%
    \label{tab:quantitative-table-accuracy-only}
\end{table*}

%% file: sections/discussion.tex
We discuss the runtime complexity and approximations of our method in general, and for invariance learning in particular, in detail in~\cref{app:complexity} and \cref{app:approximations}, respectively. 
The runtime complexity of \ourmethod, just like that of Augerino and other methods that use test-time data augmentation, increases linearly by the factor $S$, which denotes the number of augmentation samples used. 
Since we know that we are sampling from a lower bound (\cref{eq:lower_bound_loglik}), more samples are generally better and are expected to improve the performance, which we also observed in our experiments. 
In addition, \ourmethod requires estimation and differentiation of the log-determinant term. 
While our extension of \kfac to sampling-based invariant models and the batched gradients make such computation at all tractable, it can still be expensive for a large number of classes $C$ and augmentation samples $S$.
Using the \ef instead of the \ggn overcomes scaling in $C$ but is a cruder approximation.
Although these additional computations make \ourmethod slower than Augerino, they enable overcoming its issues, i.e., parameterisation-dependence and additional hyperparameters that need to be tuned~(\cref{app:augerion-comparison}), enable to learn soft invariances much more preciselly.

%% file: sections/conclusion.tex
We presented a method that enables automatic invariance learning in deep neural networks directly from training data, without requiring supervision or validation data. The approach is inspired by using the marginal likelihood, which is a parameterisation-independent quantity coinciding with generalisation performance. To make this practical, we use a differentiable Laplace approximation to allow for gradient-based optimisation of invariances in deep learning. While the accuracy of the approximation is difficult to verify, we do show experimentally that the method is capable of learning invariances in MNIST, FashionMNIST, and CIFAR-10 datasets, leading to better marginal likelihoods and higher test performances.
Our work shows that approximate Bayesian inference methods can be useful for learning complex hyperparameters, even in deep learning, and are therefore relevant beyond predictive uncertainty estimation.
In future work, it would be interesting to improve the scalability and accuracy of marginal likelihood approximations which could enable learning even more complex hyperparameters, such as augmentation distributions parameterised by neural networks.
Alternatively, improving parameterisations of invariances in neural networks could greatly improve the scalability of \ourmethod and related approaches.

%% file: sections/appendices/technical_details.tex
We apply the re-parameterization trick \citep{kingma2013auto} to define the augmentation distribution as a learnable probability distribution that is differentiable with respect to the parameters:
\begin{equation}
\begin{split}
z \in p(\vxaug \given \vx, \augmentation) \to 
z = g_{\boldsymbol{\epsilon}}(\vxaug \given \vx, \augmentation)
\end{split}, \hspace{1cm}
\begin{split}
\boldsymbol{\epsilon} \in U[-1, 1]^k\,.
\end{split}
\end{equation}

\noindent
A general affine parameterization \citep{benton2020learning} can be obtained using $k\text{=}6$ generator matrices $\mG_1, \ldots, \mG_6$ and learnable parameters $\boldsymbol{\eta} = [\eta_1, \cdots, \eta_6]^T$ for respective horizontal translation, vertical translations, rotations, horizontal scaling, vertical scaling, and shearing:

\begin{align*}
\begin{split}
\mG_1 &= 
\begin{bmatrix}
\ 0 & \ 0 & \ 1 \\
\ 0 & \ 0 & \ 0 \\
\ 0 & \ 0 & \ 0
\end{bmatrix}
\end{split}
, \hspace{0.5cm}
\begin{split}
\mG_2 &= 
\begin{bmatrix}
\ 0 & \ 0 & \ 0 \\
\ 0 & \ 0 & \ 1 \\
\ 0 & \ 0 & \ 0
\end{bmatrix}
\end{split}
, \hspace{1.5cm}
\begin{split}
\mG_3 &= 
\begin{bmatrix}
\ 0 & -1 &  \ 0 \\
\ 1 & \ 0 & \ 0 \\
\ 0 & \ 0 & \ 0
\end{bmatrix}
\end{split}
,
\\
\\
\begin{split}
\mG_4 &= 
\begin{bmatrix}
\ 1 & \ 0 & \ 0 \\
\ 0 & \ 0 & \ 0 \\
\ 0 & \ 0 & \ 0
\end{bmatrix}
\end{split}
, \hspace{1.5cm}
\begin{split}
\mG_5 &= 
\begin{bmatrix}
\ 0 & \ 0 & \ 0 \\
\ 0 & \ 1 & \ 0 \\
\ 0 & \ 0 & \ 0
\end{bmatrix}
\end{split}
, \hspace{1.5cm}
\begin{split}
\mG_6 &= 
\begin{bmatrix}
\ 0 & \ 1 & \ 0 \\
\ 1 & \ 0 & \ 0 \\
\ 0 & \ 0 & \ 0
\end{bmatrix}
\end{split}
.
\end{align*}

\noindent
To calculate $\vx'_{(x', y')}$, defined to be the value of $\vx'$ corresponding location $(x', y') \in \R^2$ on the 2-dimensional grid, we apply an inverse of the forward transformation matrix to find pixel locations in the original image:
\begin{equation}
\begin{bmatrix}
x \\ y
\end{bmatrix}
=
\exp \left( \sum_i \epsilon_i \eta_i \mG_i \right)^{-1}
\begin{bmatrix}
x' \\ y' 
\end{bmatrix}
\end{equation}
where $\exp(M) = \sum_{n=0}^\infty \frac{1}{n!}M^n$ denotes the matrix exponential \citep{moler2003nineteen}, and the transformations becomes
\begin{equation}
g_{\boldsymbol{\epsilon}}(\vx_{(x', y')}'|\vx, \augmentation) = \vx_{(x, y)}
\end{equation}
where pixel values in the output $\vx'_{(x', y')}$ are calculated exactly on the grid, and the locations in the input $\vx_{(x, y)}$ are obtained through bilinear sampling, which can be, as all of the other steps, automatically differentiated \cite{jaderberg2015spatial}. Furthermore, the entire process is highly efficient as the matrix exponential and inverse are applied on a very small 3x3 matrix and the grid resampling steps are fully parallelizable across all pixels.

%% file: sections/appendices/augerino_comparison.tex
There are two main failure modes of Augerino that our method overcomes.

\begin{figure}[h]
\vspace{-1em}
    \centering
    \resizebox{\linewidth}{!}{
    \includegraphics{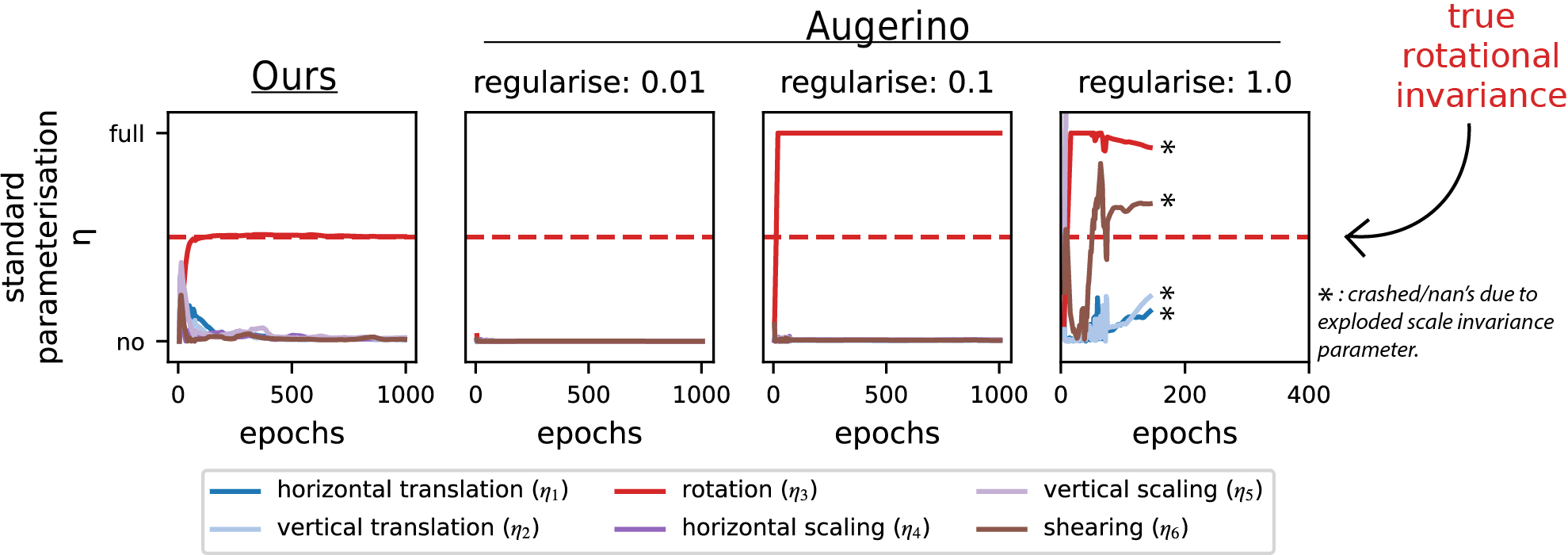}
    }
    \caption{Learning invariance on partially-rotated MNIST with different settings of Augerino regularisation strength.\vspace{-1em}}
    \label{fig:augerino_failure_tuning}
\end{figure}

\paragraph{Failure case 1: Augerino requires additional hyperparameter that needs tuning.}
The regularisation used in Augerino introduces an additional hyperparameter that needs tuning. In our experiments, we find that this hyperparameter is non-trivial to tune and requires an additional validation set. Since our method follows from Bayesian model selection, there is no additional hyperparameter. In \cref{fig:augerino_failure_tuning}, we train Augerino on partially-rotated MNIST data using different settings for regulariser hyperparameter $\{0.01, 0.1, 1.0\}$ and compare with our model. We observe that Augerino has difficulty learning partial invariance and is dependent upon the setting of the hyperparameter. Our method, on the other hand, does learn partial rotational invariance without having to tune an additional hyperparameter.

\paragraph{Failure case 2: Augerino depends on the used parameterisation of invariance.}
Augerino depends on the parameterisation of the invariance parameter as it uses the heuristic that increasing this parameter corresponds to increased invariance. This makes it fail, for example, under a change of variables $\eta \mapsto \frac{1}{\eta}$. Our method can be applied without requiring knowledge of the particular parameterisation. We demonstrate this failure case in \cref{fig:augerino_failure_parameterisation} below, by training our method (left) and Augerino (right) using standard parameterisation $\eta$ (top) and inverted parameterisation $\frac{1}{\eta}$ (bottom). Unlike Augerino, our method learns the correct (in this case rotational) invariance independent of the used parameterisation.
\begin{figure}[h]
\vspace{-1em}
    \centering
    \resizebox{0.5\linewidth}{!}{
    \includegraphics{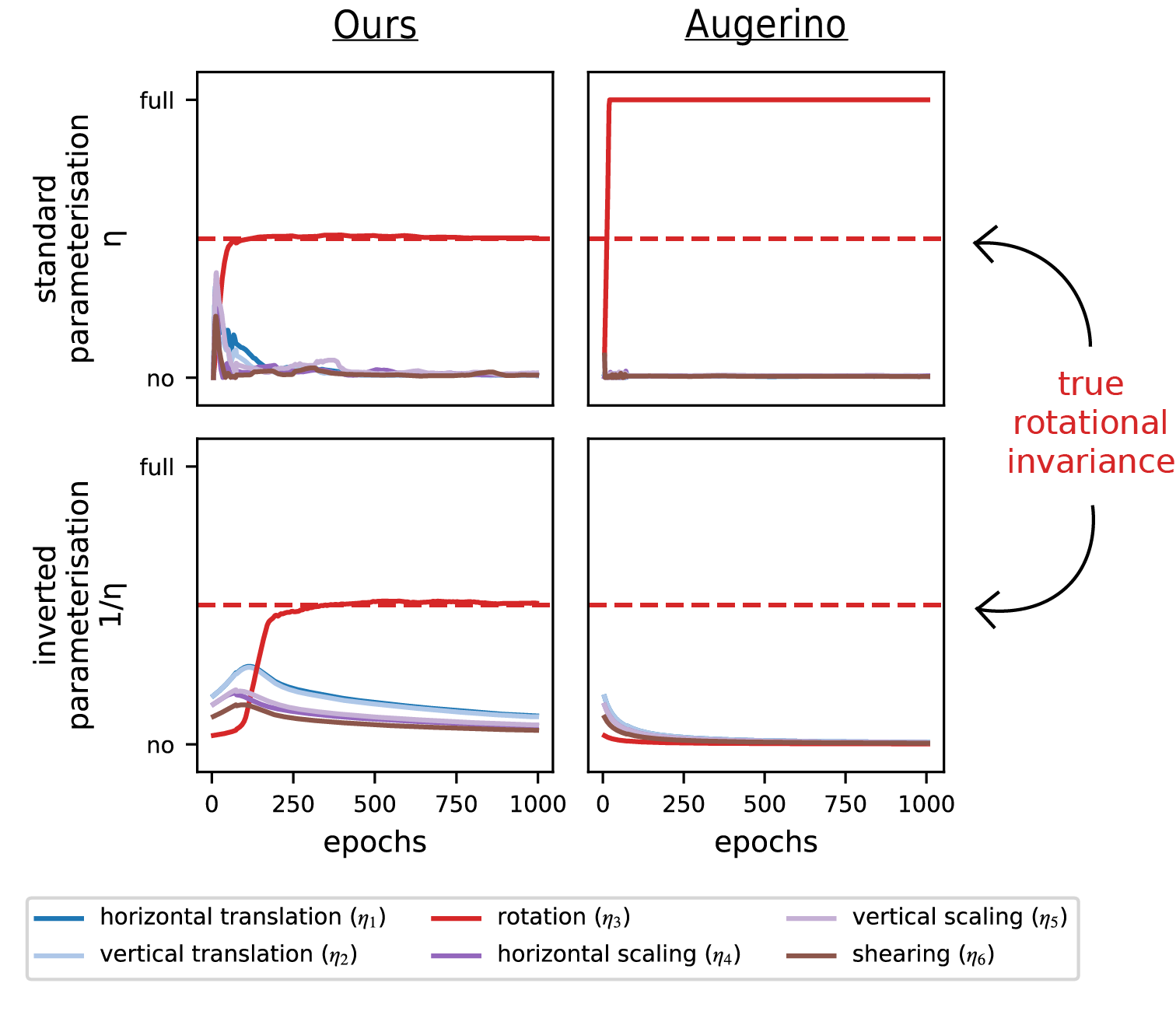}
    }
    \caption{Learned invariance on partially-rotated MNIST for different invariance parameterisations: standard parameterisation $\eta$ (top) and inverted parameterisation $\frac{1}{\eta}$ (bottom). Comparison between our invariance learning approach (left) and Augerino (right). Unlike Augerino, our method is able to learn the correct (in this case rotational) invariance independent of the used parameterisation.}
\vspace{-1em}
    \label{fig:augerino_failure_parameterisation}
\end{figure}

%% file: sections/appendices/complexity.tex
In this section, we discuss the computational and memory complexities of the proposed method.
In particular, we review the complexity of the \ggn approximation, discuss the complexity of the \ggn for invariant neural networks, and give the total complexity of \emph{estimating} and \emph{differentiating} the Laplace-\ggn approximation using \kfac to the log marginal likelihood, which is necessary for \ourmethod.

\subsection{The complexity of Vanilla \ggn Approximations}
When not dealing with invariant neural networks that use sampling, the complexities are well-known for diagonal, \kfac, and a full \ggn approximation. 
We have $N$ data points, $C$ neural network outputs, e.g. classes, and $P$ neural network parameters.
In general, we assume that $C \ll P$ and also that $C \ll G$, where $G$ is the size of hidden representations.

The full \ggn approximation in \cref{eq:ggn} is in \order{NP^2C} for computing $N$ matrix-products. Computing its log-determinant additionally costs \order{P^3}.
The diagonal \ggn approximation would be in \order{NPC} and computation of the log-determinant only \order{P}.
The complexity of \kfac-\ggn depends on the type of layer and its in- and output dimensionalities D and G, respectively, as defined in \cref{sec:kfac_aug}.
Then, the complexity of computing \cref{eq:kfac_general} is \order{ND^2+NCG^2}. 
The first term is due to summation of $N$ matrices, each of which is an outer product of $D$-dimensional vectors. 
The second term is due to the outer product with $\mLambda_n \in \R^{C \times C}$ in between and since we assume $C \ll G$, we have $CG^2$ at worst for one such product.
Computing the log-determinant can be done efficiently in \order{D^3+G^3} by decomposing the Kronecker factors~\citep{immer2021scalable}.
For typical neural networks, both $D$ and $G$ are below $1000$ and computing the log-determinant is tractable.

\subsection{The Effect of Sampling in Invariant Neural Networks on the Complexity}
Our invariant formulation of neural networks through perturbations requires $S$ approximation samples.
These samples also make computation of the \ggn approximations more expensive.
The computation of the log determinants remains the same as the sizes of the \ggn approximations remain equivalent.

The complexity of a full \ggn computation changes to \order{NP^2C + NPCS} where the first term is due to the Jacobian products as above and the second term is for computing and averaging $S$ Jacobians of size $P\times C$ over all $N$ data points. 
The same term without augmentation samples would be \order{NPC} and is therefore contained in the above complexity \order{NP^2C}.
Because we usually have $S \ll P$, the complexity of augmented full \ggn (\cref{eq:aug_ggn}) is also unchanged \order{NP^2C}.

The complexity of the diagonal-\ggn approximation increases relatively worse than the full \ggn as it goes from \order{NPC} to \order{NPCS}.
This is because the diagonal-\ggn still requires computation of the $NS$ Jacobians of size $PC$.
In total, the computation of the diagonal approximation therefore scales linearly in the number of samples $S$.

The complexity of \kfac-\ggn with augmentation samples changes to \order{N(D^2 + CG^2 + DS + CGS)} using our proposed method in \cref{eq:kfac_aug}.\footnote{If we don't use the proposed additional approximation, the complexity would grow too large to be tractable since the Kronecker-factored structure would not be maintained but lead to a dense Matrix of size $DG \times DG$.}
The last two terms dependent on $S$ come up due to the aggregation of augmentation samples in our approximation, that is, the expectations over $\va$ and $\vg$ in the second line of \cref{eq:kfac_aug}.
However, we typically have that $S \leq D$ and $s \leq G$ and therefore the computational complexity is still in \order{N(D^2 + CG^2)}.
For typical settings, the computational complexity of the \kfac-\ggn therefore does not increase over the vanilla variant.
This makes the proposed \kfac approximation the only option for invariance learning with differentiable Laplace approximations among the considered ones since the diagonal scales unfavourably in $S$ and the full \ggn is too expensive.
In principle, one could apply the same approximation as in \cref{eq:inv_kfac_jac} for the diagonal \ggn but since it even performs worse than the \kfac variant without further approximations~(cf.~\cref{app:approximations}), \kfac is preferred.

If we assume that $D=G$, that is, the in and output size of the layer of interest are equivalent, and we only consider this single fully-connected layer to carry all parameters $P=DG=D^2=G^2$, it becomes simple to compare the complexities of all three \ggn estimators with and without augmentation. This setting is not unrealistic for fully-connected layers where hidden sizes remain rather similar in the middle of the network.
The complexities simplify as shown in the table below.
\begin{center}
\begin{tabular}{c|c c c}
    & full & \kfac & diagonal \\
    \midrule
    vanilla \ggn & \order{NP^2C} & \order{N(D^2+CG^2)} = \order{NPC} & \order{NPC} \\
    augmented \ggn & \order{NP^2C} & \order{N(D^2+CG^2+DS+CGS)} = \textcolor{red}{\order{NPC}} & \textcolor{red}{\order{NPCS}} \\
    (log)det & \order{P^3} & \order{G^3+D^3} = \order{P^{1.5}} & \order{P} \\
\end{tabular}
\end{center}

In this setting the vanilla \kfac-\ggn approximation is as cheap to compute as the diagonal approximation and only the computation of the log-determinant is slightly more complex.
Even more relevant to us, though, is that the computation of the augmented variant of \kfac has the same computational complexity as the vanilla variant and an $S$ times lower complexity than the diagonal augmented \ggn, as marked in \textcolor{red}{red} in the table.
This is the case although the \kfac variant works better than the diagonal variant~(see. \cref{app:approximations}).
The vanilla \kfac variant often achieves performance on par with the full \ggn despite the greatly reduced computational complexity and a much better performance than the diagonal variant~\citep{ritter2018scalable, immer2021scalable, immer2020improving, daxberger2021laplace}.

\subsection{Computational Complexity of Laplace-\ggn Estimation and Differentiation}
Above, we discussed the computational complexity of the log determinant component of the Laplace-\ggn~(\cref{eq:aug_laplace_ggn}).
The total computational complexity of the Laplace-\ggn also depends on the first term, the log likelihood loss, which is independent of the \ggn approximation itself with a computational complexity of \order{NPS} and memory complexity \order{MPS} for differentiation since it allows for stochastic batching in $N$.
This gives us the total computational complexity for estimating the Laplace-\ggn marginal likelihood by summing up the cost of the log likelihood loss (\order{NPS}), the cost of the augmented \ggn as in the above table, and the cost of computing its log determinant.
For our proposed \kfac variant, this gives a total computational complexity of \order{NPS+NPC+P^{1.5}} for the simplified architecture assumptions.
Naive automatic differentiation would result in the same memory cost for its differentiation, which is intractable.
In the next section, we propose a method that allows batching all computation and reducing the memory complexity of differentiating the \ggn (or \kfac) to \order{MPC} where $M$ is a free parameter and can be as small as $1$ for a single data point, resulting in a total memory complexity of \order{MPS+MPC}, where the first term comes from the batch-friendly log likelihood loss.
This effectively makes our method possible by reducing the required memory from terabytes to gigabytes in larger scale examples.

\subsection{Measured Computation Time for Different Methods}

To give an idea of computation time in practice, we measure parameter and hyperparameter updates on an NVIDIA A30. We report times for training a ResNet (8-8) on CIFAR-10, a CNN on F-MNIST and an MLP on MNIST. We find that parameter updates using maximum likelihood training take 0.16 seconds, 0.08 seconds or 0.06 seconds per epoch, on the respective datasets. When using $S{=}11$ samples on Augerino and \ourmethod, this becomes 1.9 seconds, 0.5 seconds and 0.3 seconds for the parameter updates in both methods. This is in line with the linear complexity increase of the models in $\mathcal{O}(S)$. The additional hyperparameter update of \ourmethod estimating the marginal likelihood and differentiating it takes 34.5 , 11.1 and 12.9 seconds, for the respective datasets. For the \ef approximation, it takes 14.4 , 7.3, and 5.9 seconds, respectively. The improved hyperparameter gradient of \ourmethod increases the cost over Augerino by roughly a factor of 10 in these empirical timings.
However, the theoretical analysis above shows that \ourmethod has the same asymptotic complexity as Augerino for $C \leq S$ and $\sqrt{P} \leq NS$, which commonly holds. We hypothesise that a more efficient implementation could therefore further benefit empirical runtimes.

%% file: sections/appendices/jvp.tex
In \cref{sec:grad_aug}, we discussed how to efficiently backpropagate through the log-determinant of the Laplace-\ggn objective in \cref{eq:aug_laplace_ggn}.
The log-determinant does not allow for batching and straightforward backpropagation because the \emph{memory complexity} of the log-determinant backpropagation is too high as it would be equivalent to the \emph{total computational complexity} of the \ggn approximation.
In the comparison of complexities (\cref{app:complexity}), the total computational complexity would be the sum of computing the \ggn approximation and the respective log-determinant computation.
Following the example in \cref{app:complexity}, using our \kfac-\ggn approximation, we would have \order{NPC + P^{1.5}} \emph{memory complexity} to compute the gradient with respect to augmentation parameters $\augmentation$, where the first term is clearly intractable as it is the product of data points $N$ and parameters $P$, even though we are considering a simplified case with $S \ll D = G$ for \kfac and it would normally even scale in the number of samples $S$.
The terms other than the log-determinant are simpler to handle, and can be computed using backpropagation:
the first term, the conditional log likelihood, can be batched over the $N$ data points due to the sum.
The second term, the log prior of parameters $\vtheta$, is independent of the augmentation parameters.

Our approach relies on computing a \emph{preconditioner} first that incurs a low memory complexity and can be computed in aggregation and then using this result to compute a stochastic, but unbiased, gradient.
The preconditioner acts as the vector in a vector-Jacobian product~\citep{paszke2017automatic}.
This method allows to decouple computational complexity from memory complexity and is \emph{necessary} to enable learning invariances with the marginal likelihood.
For \kfac the computational complexity will be unchanged \order{NPC+P^{1.5}} and the memory complexity is \order{MPC} with batch size $M$ as opposed to the intractable \order{NPC+P^{1.5}}.
In \cref{sec:grad_aug}, we proposed the method in general for full and diagonal \ggn log determinants.
However, this method does not directly extend to \kfac since it would require evaluation of an expensive Kronecker product.
Here, we apply the same idea to show how gradient accumulation is possible for \kfac log determinants.

\subsection{Derivation for \kfac}
\label{app:grad_kfac}
If we apply the method above directly to \kfac, we would have to compute $\mH\inv$ for a block corresponding to a single layer which is in most cases too large as it breaks the Kronecker-factored structure.
In particular, $\mH^{\kfac} = (\mA \otimes \mG + \delta \mI)$ is a typical form that the Kronecker-factored block would take.
However, it is inefficient to compute and store such large matrix that is quadratic in the size of the number of parameters of the respective layer.
Instead, we would like to maintain the Kronecker-factored structure.
Since we need to compute the eigendecomposition of $\mA$ and $\mG$ to efficiently estimate the marginal likelihood approximation~\citep{immer2021scalable}, we can make use of it as preconditioner for batched gradient estimation.
In fact, it is straightforward to start from eigenvalues $\vlambda \in \R^P$ of $\mH \in \R^{P \times P}$ instead of $\mH$ itself since these can be used to simply compute the log-determinant as a sum of the log of eigenvalues:
\begin{equation}
    \log \vert \mH \vert = \log \prod_{p \in [P]} \lambda_p = \sum_{p \in [P]} \log \lambda_p.
\end{equation}
For a single block of the Kronecker-factorization and ignoring the normalization by $\tfrac{1}{N}$ for notational convenience, we have $\mH \approx \mA \kron \mG + \delta \mI$~(\cref{eq:kfac_general}) and therefore 
\begin{equation}
    \log \vert \mH \vert = \log \vert \mA \kron \mG + \delta \mI \vert = \sum_{a \in [D],g \in [G]} \log (\lambda_{a} \lambda_{g} + \delta),
    \label{eq:kfac_logdet_eig}
\end{equation}
where $\lambda_a$ denotes the $a$th eigenvalue of $\mA$ and $[D]=1\ldots D$ are the dimensions of $\mA$~(cf.~\cref{sec:background}), and identically for $\lambda_g$.
We have a further dependency of $\lambda_a$ and $\lambda_g$ values on the hyperparameter $\eta \in \R$, i.e. $\lambda_a(\eta)$ and $\lambda_g(\eta)$, which is implicit for notational simplicity.
Here, we only consider a scalar hyperparameter $\eta$ but the computation, and in particular, the final vector-Jacobian product extend to the vector case.
The derivative w.r.t.~$\eta$ is given by
\begin{align}
    \diffaug \sum_{a,g} \log (\lambda_{a} \lambda_{g} + \delta)
    &= \sum_{a,g} (\lambda_{a} \lambda_{g} + \delta)\inv \diffaug (\lambda_{a} \lambda_{g}) \nonumber \\
    &= \sum_{a,g} (\lambda_{a} \lambda_{g} + \delta)\inv [(\diffaug \lambda_{a}) \lambda_{g} + \lambda_a (\diffaug \lambda_g)] \label{eq:kfac_eig_grad} \\
    &= \sum_{a} (\diffaug \lambda_a) \sum_g \lambda_g (\lambda_{a} \lambda_{g} + \delta)\inv 
     + \sum_{g} (\diffaug \lambda_g) \sum_a \lambda_a (\lambda_{a} \lambda_{g} + \delta)\inv. \nonumber
\end{align}
The two summands in the last expression require the same operations for both $\mA$ and $\mG$. 
We will only derive the final expression for the first one, the second one follows accordingly after swapping indices $g$ and $a$.
To further simplify, we need the gradient of an eigenvalue $\lambda_a$ with respect to the augmentation.
Since $\lambda_a$ is an eigenvalue of $\mA$, we will use the chain-rule and have $\diffaug \lambda_a = \vv_a\transpose [\diffaug \mA] \vv_a$ with $\vv_a \in \R^D$ as the $a$th eigenvector of $\mA$ corresponding the eigenvalue $\lambda_a$.
We can then continue to simplify as follows:
\begin{align}
    \sum_{a} (\diffaug \lambda_a) \sum_g \lambda_g (\lambda_{a} \lambda_{g} + \delta)\inv  
    &= \sum_{a} (\vv_a\transpose [\diffaug \mA] \vv_a) \sum_g \lambda_g (\lambda_{a} \lambda_{g} + \delta)\inv \nonumber\\
    &= \sum_{i,j \in [D]} \Big[ \sum_{a} \vv_{a,i} \vv_{a,j} \underbrace{\sum_g \lambda_g (\lambda_{a} \lambda_{g} + \delta)\inv}_{\defeq c_a} \Big] \Big[\diffaug \mA_{i,j}\Big] \\
    &= \mathrm{vec}(\sum_{a} c_a \vv_a \vv_a\transpose)\transpose [{\textstyle \sum_{n=1}^N} \diffaug \mathrm{vec}(\mA_n(\eta))], \nonumber
\end{align}
where the last line can again be expressed as a Jacobian-vector product and the second term sums over the Kronecker factor $\mA_n$ per data point $n$ and can be batched or estimated stochastically.
In particular, the vector that depends on the eigenvalues $\lambda_\cdot$ and eigenvectors $\vv_a$ can be computed after a full dataset pass using a single eigendecomposition of the Kronecker factors.
Then, the second term, which requires computing the gradient with respect to the augmentation parameters, can be estimated with low memory footprint by batching and/or a stochastic estimate of the second term.

%% file: sections/appendices/approximations.tex
We discuss the approximations necessary for the proposed approach to invariance learning with Bayesian model selection.
The discussion extends to other complex hyperparameters that could be optimised using the methodology described.
Our algorithm is motivated by Bayesian model selection with an empirical Bayes procedure, where we optimise the hyperparameters $\augmentation$ according to their maximum likelihood on the second level of inference (ML-II).
Since the necessary marginal likelihood, $p(\data \given \augmentation)$, in \cref{eq:marglik} is intractable for neural networks, our approach relies on several approximations detailed below in \cref{app:approx_list}.
Despite the approximations, our experiments show that the proposed method is able to recover invariances present in the data without any supervision, which empirically validates the approach.
To that end, we show in \cref{app:approx_comparison} that the proposed \kfac approximation is necessary to reliably recover invariances while cheaper approximations like the empirical Fisher or diagonal \ggn are insufficient.

\subsection{List of Approximations to the Log Marginal Likelihood for an Invariant Model}
\label{app:approx_list}

\paragraph{Laplace-\ggn Approximation.}
Instead of a vanilla Laplace approximation that suffers from estimation issues due to the full-network Hessian as well as potential indefiniteness of it, we use the Laplace-\ggn to approximate the log marginal likelihood.
The Laplace-\ggn~\citep{khan2019approximate} is more efficient and stable to estimate and has a clear justification due to its equivalence to linearised Laplace~\citep{foong2019between, immer2020improving, antoran2022adapting}.
The first step of Laplace-\ggn is to linearise a neural network around an arbitrary linearisation point $\paramstar$, i.e., $\vf^{\textrm{lin}}_\paramstar(\vx;\param) = \vf(\vx;\paramstar) + \mJ(\vx;\paramstar) (\param - \paramstar)$.
The error of this Taylor approximation is then in the order of the second derivative of $\vf(\vx;\param)$ at $\paramstar$.
\citet{immer2020improving} argue that this changes the model by modifying the underlying likelihood function to depend on $\vf^{\textrm{lin}}$ instead of $\vf$ and inference takes place in the first order term $\mJ(\vx;\paramstar) \param$, i.e., we have $p(\vy \given \vf^\textrm{lin}_{\paramstar}(\vx;\param), \hypothesis)$ and an unchanged prior.
We can therefore understand such approximation as a modification of the model itself.
We arrive at the Laplace-\ggn approximation by applying the Laplace approximation to the linearised model.
Because we deal with a linear(ised) model, the Laplace approximation tends to perform well with large amounts of data in the case of classification~\citep{bishop2006pattern} and is exact for a Gaussian likelihood in the case of regression~\citep{foong2019between}.
In the large data limit, the Laplace approximation itself becomes asymptotically exact~\citep{dehaene2019deterministic}.
Linearisation becomes asymptotically exact in the infinite width limit of neural networks, which results in the neural tangent kernel~\citep{jacot2018neural}.

\paragraph{Sampling augmentations.}
The \ggn of an invariant model follows from the linearisation of such model~(see above paragraph).
The only additional approximation arises due to the sampling necessary to approximate the expectation over $p(\vxaug \given \vx, \augmentation)$ as in \cref{eq:aug_model_sampled}.
Using Jensen's inequality it can be shown that a Monte Carlo approximation of the invariant neural network in the log likelihood leads to a lower bound of it. 
In particular, both \citet{nabarro2021data} and \citet{schwoebel2022layer} show
\begin{align}
    \log p(\vy \given \vfaug(\vx;\param, \augmentation)) 
    &= \log p(\vy \given \E_{\vx'_1, \ldots, \vx'_S} [\tfrac{1}{S} \textstyle \sum_s \vf(\vx'_s;\param)]) \nonumber \\
    &\geq \E_{\vx'_1, \ldots, \vx'_S} [\log p(\vy \given \tfrac{1}{S} \textstyle \sum_s \vf(\vx'_s;\param))], \label{eq:lower_bound_loglik}
\end{align}
where the lower bound is simply due to Jensen's and the fact that we use minimal exponential family likelihoods such that $\vf$ is the natural parameter~(cf.~\cref{sec:background}).
In practice, we approximate the last term by sampling one set of augmented inputs $\vx'_1,\ldots, \vx'_S$.
The more samples $S$ we take, the tighter the bound. 
Therefore, it is generally desired to use as many samples as affordable.

\paragraph{Augmented \kfac approximation.}
\kfac of an invariant neural network relies on further approximations to the intractable \ggn matrix that is quadratic in the number of neural network parameters.
We first discuss the approximations of vanilla \kfac before the additional approximation that is necessary for invariant neural networks.
\kfac approximates the full $P\times P$ \ggn by a block-diagonal approximation that captures the \ggn over each layer independently in form of a block-diagonal matrix.
Further, to maintain an efficient Kronecker-factored structure that is only exact for a single data point~(cf.~\cref{eq:kfac_ggn_start}) a product of sums is approximated by sums of products~(cf.~\cref{eq:kfac_general}).
In practice, this approximation has been validated in the context of optimisation~\citep{martens2015optimizing, osawa2020scalable, dangel2019backpack}, posterior approximation~\citep{ritter2018scalable, zhang2018noisy, osawa2019practical, immer2020improving}, and model selection~\citep{immer2021scalable, immer2022probing}. 
Both approximations are only exact when the model is linear.
The additional approximation proposed by us applies to the augmentation samples and is similar to the second approximation of vanilla \kfac but applied to the Jacobians of the invariant neural network in~\cref{eq:inv_kfac_jac}.
In our experiments, we empirically verify that such approximation does not lead to issues.
Below in \cref{app:approx_comparison}, we show that this approximation still leads to better performance than a diagonal \ggn that avoids it.
For an invariant linear model, the proposed \kfac extension would be exact as was the case for vanilla \kfac:
\begin{align}
    \hat{\mJ}_{\param}(\vx_n; \augmentation) = \E_{p(\vxaug_n | \vx_n, \augmentation)} [ \vxaug_n \kron \mI] = \E_{p(\vxaug_n | \vx_n, \augmentation)} [\vxaug_n] \kron \mI
\end{align}
gives the Jacobian of an invariant linear model and requires no further approximation since the product separates naturally. 
This is also the case for the last layer of a neural network.

\subsection{Diagonal and Empirical Fisher Approximations}
\label{app:approx_comparison}
\begin{figure}[t]
    \centering
    \includegraphics{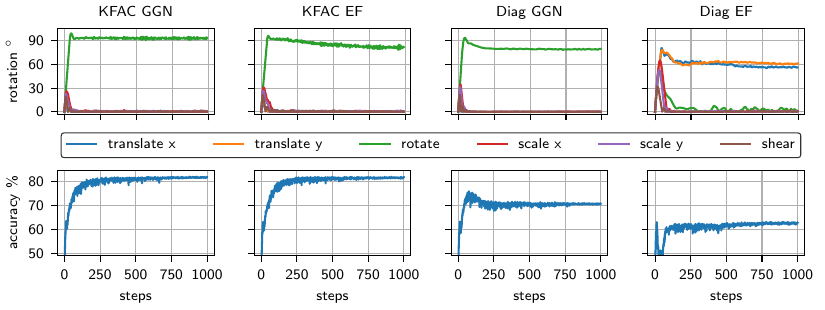}   
    \vspace{-1em}
    \caption{Comparison of \kfac and diagonal \ggn and \ef approximations for invariance learning using the corresponding log marginal likelihood approximation on $5000$ random $\pm90^\circ$ rotated FashionMNIST samples.
    \kfac performs significantly better than diagonal approximations.
    \ggn is slightly better than \ef for \kfac but significantly better in the diagonal case.} 
    \label{fig:approx_ablation}
\end{figure}

Here, we briefly discuss the reason for extending the \kfac-\ggn approximation to invariant neural networks instead of using a diagonal approximation.
The argument is both computational, following from \cref{app:complexity}, and empirical, following from the better performance of \kfac than diagonal.
We further discuss the empirical Fisher, which is an alternative to the \ggn but has a $C$ times lower computational complexity where $C$ is the number of outputs~(classes).
The empirical Fisher is a viable alternative but is not justifiable by a linearisation like the \ggn and is known to have certain pathologies~\citep{kunstner2019limitations}.
\cref{fig:approx_ablation} shows the invariance learning performances for partially rotated ($\pm 90^\circ$) FashionMNIST with \kfac and diagonal each with \ggn and empirical Fisher variant.
The setup is identical to the one used in the experiments and is conducted on $5000$ randomly chosen data points over $3$ seeds.
The figure only contains the run of a single seed but the observation is consistent across all runs.

\paragraph{Diagonal approximations.}
Although it does not model any correlations, a diagonal \ggn approximation is typically only marginally more efficient than a \kfac approximation~(cf.~\cref{app:complexity} for a detailed discussion of complexities).
This is due to the efficient approximations employed in \kfac that could potentially lead to a worse overall performance but empirically it has been found that \kfac performs always at least as well as a diagonal approximation as discussed in \cref{app:approx_list}.
When considering invariance learning with \ggn approximations, the proposed \kfac for invariant neural networks further significantly reduces the runtime and memory complexity in the augmentation samples $S$, which we discuss in \cref{app:complexity}.
\cref{fig:approx_ablation} further shows that \kfac performs significantly better in terms of test accuracy and in terms of recovering the underlying $\pm 90^{\circ}$ rotational invariance.
Overall, this leaves no reason to use a diagonal approximation for learning complex hyperparameters.

\paragraph{Empirical Fisher.}
The empirical Fisher is an approximation to the true Fisher, which is in turn equivalent to the \ggn for the cases we consider, but can be significantly cheaper to compute.
For example in the case of classification with $C$ classes, the cost of computing the \ggn scales linearly in $C$ while the empirical Fisher does not.
Mathematically, the empirical Fisher is given by summing up outer products of gradients as opposed to Jacobians.
However, the empirical Fisher does not follow from a linearisation perspective and also has certain pathologies in optimisation~\citep{kunstner2019limitations}.
In \cref{fig:approx_ablation}, we find that the empirical Fisher can work as well as the \ggn for \kfac but its diagonal approximation fails and even learns the wrong invariances.
That the diagonal empirical Fisher, the cheapest curvature approximation, fails in this setting is interesting because \citet{immer2021scalable} observe that it can reliably learn regularisation hyperparameters for ResNets.
We hypothesise that learning invariances requires better approximations and profits from approximation quality.
The \kfac empirical Fisher can be extended to invariance learning similar to the \ggn and requires averaging the gradients, which are used for the outer products, over augmentation samples.
In \cref{tab:large-resnet-results}, we additionally compare its performance for a wide ResNet on standard CIFAR-10, where it performs almost as well but with $10$-fold speed-up.

%% file: sections/appendices/mechanism.tex
We discuss the mechanism by which the Laplace approximation to the marginal likelihood enables invariance learning and why simple maximum likelihood is insufficient.
Leaving out terms that do not depend on the invariance parameter \augmentation, such as the prior $\log p(\paramstar)$, the Laplace-\ggn approximation to the log marginal likelihood of an invariant neural network introduced in \cref{eq:aug_laplace_ggn} can be decomposed as
\begin{align}
  \underbrace{\underbrace{\textstyle{\sum_{n=1}^N} \log p(\vyn \given \tfrac{1}{S} {\textstyle \sum_s} \vf(\vg(\vx_n,\vepsilon_s;\augmentation);\paramstar))}_\text{maximum likelihood objective}
  - \tfrac{1}{2} \log \vert \hat{\mH}_{\paramstar}^\ggn(\augmentation) \vert}_\text{marginal likelihood objective}\, .
  \label{eq:aug_laplace_ggn_rep}
\end{align}
In the following we discuss why the maximum likelihood objective for $\augmentation$ in general will not lead to proper invariance learning while the marginal likelihood objective will.

\subsection{Regular Maximum Likelihood Does \emph{Not} Learn Invariances}
\label{app:maximum-likelihood}
\begin{wrapfigure}{r}{0.27\textwidth}
  \vspace{-3.8em}
  \begin{center}
    \includegraphics[width=0.25\textwidth]{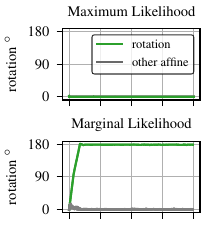}
  \end{center}
  \vspace{-3em}
\end{wrapfigure}
In the context of deep learning, the underlying non-invariant neural network function $\vf$ is complex and can arbitrarily (over-)fit the data.
Assuming such complex function, we do not need any invariance parameter to fit the data optimally according to the maximum likelihood part in above \cref{eq:aug_laplace_ggn_rep}.
In fact, changing the invariance parameter \augmentation rather hinders fitting due to sampling noise and restricting the function class. 
This behaviour is shown on fully rotated MNIST in the figure on the right where all affine invariance parameters, also rotation, remain zero.

\subsection{Marginal Likelihood Learns Invariances}
The marginal likelihood objective arises from the principle of Bayesian model selection, which trades off performance and model complexity and is introduced in common machine learning literature~(\citealp[Sec. 28]{mackay2003information}; \citealp[Sec. 3.4 and 4.4]{bishop2006pattern}; \citealp[Sec. 5.6]{murphy2012machine}) and briefly in \cref{sec:background}.
\emph{Intuitively}, a model is less complex if it can explain two data points $\vx, \vx'$, where one is a transformed version of the other, with the same function. 
A non-invariant model needs to fit both data points individually and is therefore more complex. 
\Citet{van2018learning} first introduced invariance learning with the marginal likelihood for Gaussian processes and discuss the underlying principle of it.
In the following, we elaborate on the mechanism when using the Laplace-\ggn approximation to the log marginal likelihood for invariance learning proposed in this work.

\textit{Mathematically}, the log-determinant part of the Laplace-\ggn marginal likelihood approximation, which is the only additional term in comparison to the maximum likelihood objective, favours invariant models. 
Consider two (or more) data points $\vx_1$ and $\vx_2$ in the same orbit, i.e. one is a transformed version of the other, and we know the true invariance parameter $\augmentationstar$.
Further, we assume to have exact invariance in the data such that the perturbation distribution is uniform on the orbit of the corresponding group~(\cref{sec:background}; \citealp{kondor2018clebsch}; \citealp{ginsbourger2016}).
This means that $\vx_1$ is an augmented version of $\vx_2$ and vice versa and their perturbation distributions are identical: $\forall \vx': p(\vx' \given\ \vx_1, \augmentationstar) = p(\vx' \given \vx_2, \augmentationstar)$.
Then, an invariant model would have identical Jacobians for both data points, which leads to zero angle between Jacobians and maximizes the negative log determinant.
To see this, we use a reformulation using the matrix determinant Lemma as proposed by \citet{immer2021scalable} that allows analytic computation for two data points.
Defining $\hat{\mJ} \in \R^{2C \times P}$ that consists of the two Jacobians $\mJ_i = \hat{\mJ}_\paramstar(\vx_i;\augmentationstar) \in \R^{C \times P}$ for $i \in \{1,2\}$ and $\hat{\mLambda} \in \R^{2C \times 2C}$ that is a block-diagonal matrix with entries $\hat{\mLambda}(\vx_i;\paramstar, \augmentationstar) \in \R^{C \times C}$ for $i \in \{1,2\}$, we have
\begin{align}
 - \log \vert \hat{\mH}_{\paramstar}^\ggn(\augmentationstar) \vert 
 = - \log \vert \hat{\mJ}\transpose \hat{\mLambda} \hat{\mJ} + \delta \mI_P \vert
 \propto - \log \left[ 
 \vert \delta\inv \hat{\mJ} \hat{\mJ} \transpose + \hat{\mLambda}\inv \vert \, \vert \hat{\mLambda} \vert
 \right],
\end{align}
where $\delta$ denotes the prior precision of a Gaussian prior.
For simplicity, assume a Gaussian likelihood with observation noise $\sigma=1$, prior $\delta=1$, and a single output $C=1$.
Then, we have 
\begin{align}
 - \log \vert \hat{\mH}_{\paramstar}^\ggn(\augmentationstar) \vert 
 \propto 
 &- \log \vert \hat{\mJ} \hat{\mJ}\transpose + \mI_2 \vert
 =
 - \log \left( 1 + \|\mJ_1\|_2^2 + \| \mJ_2 \|_2^2 + \| \mJ_1 \|_2^2 \| \mJ_2 \|_2^2 - (\mJ_1 \mJ_2\transpose)^2  \right) \nonumber \\
 &= - \log \left( 1 + \|\mJ_1\|_2^2 + \| \mJ_2 \|_2^2 + (1 - \cos^2 \phi) \|\mJ_1\|_2^2 \| \mJ_2 \|_2^2
 \right), 
\end{align}
where $\phi$ is the angle between Jacobians $\mJ_1, \mJ_2$ of $\vx_1$ and $\vx_2$, respectively.
Therefore assuming fixed Jacobian norms, the negative log determinant is maximized when we have $\cos^2 \phi = 1$, i.e., when Jacobians have zero angle $\phi = 0$ modulo direction.

%% file: sections/appendices/algorithm.tex
Here, we detail the final algorithm that we proposed for invariance learning using marginal likelihood approximations.
The algorithm extends the one proposed by \citet{immer2021scalable} to complex hyperparameters.
Their algorithm is only tractable for simple hyperparameters, such as regularisation or observation noise, here denoted by $\model$.
Our contributions enable scaling to invariance parameters and other complex hyperparameters and are detailed in Lines 14 to 21 in \cref{algo}.
The algorithm below uses simple stochastic gradient updates with a fixed learning rate.
In practise, we use a decaying step size and the Adam optimiser~\citep{kingma2014adam} as detailed in the training procedure for our experiments in \cref{app:training-details}.

\begin{algorithm}
\caption{Detailed invariance learning algorithm using marginal likelihood estimates. \\
Lines 4-13 are as in \citep{immer2021scalable}, 14-21 is our extension to invariance parameters.}
\label{algo}
\begin{algorithmic}[1]
\State \textbf{Input:} dataset $\data=\{(\vx_n,\vy_n)\}_{n=1}^N$, simple hyperparams (observation and prior noise) $\model$, invariance parameter $\augmentation$, likelihood $p(\data \given \param, \augmentation, \model)$, prior $p(\param \given \model)$, number of MC samples $S$, batch size $M$, number of epochs $T$, neural network learning rate $\gamma_{\param}$, hyperparamater learning rate $\gamma_{\model}$, invariance parameter learning rate $\gamma_{\augmentation}$
\State \textbf{Initialise: } neural network parameters $\param$ (e.g.~\citet{he2016deep} init.), \model and $\augmentation = \vzero$
\For{$T$ epochs}
  \For{each batch $\sB \subseteq \data$ of size $|\sB| \leq M$} \Comment{SGD training of neural network}
    \State $\param \gets \param + \gamma_\param \nabla_\param \tfrac{N}{|\sB|}\sum_{m \in \sB} \log p(\vy_m \given \tfrac{1}{S} \sum_{s=1}^S \vf(\vg(\vx_n,\vepsilon_s;\augmentation);\param), \model) + \log p(\param \given \model)$
  \EndFor
  \State $\hat{\mH}_{\param}^\ggn \gets \vzero$
  \For{each batch $\sB \subseteq \data$ of size $|\sB| \leq M$} \Comment{Compute $\vfaug$ and \ggn without comp. graph}
    \State $\vfaug_{m \in \sB} \gets [\tfrac{1}{S} \sum_{s=1}^S \vf(\vg(\vx_m,\vepsilon_s;\augmentation);\param) ]_{m \in \sB}$ 
    \State $\hat{\mH}_{\param}^\ggn \gets \hat{\mH}_{\param}^\ggn + \sum_{m \in \sB} \hat{\mH}_m^\ggn$ \Comment{Typically use \kfac}
  \EndFor
  \State $\hat{\mH}_\param(\model) \gets \alpha_{\model} \hat{\mH}_\param^\ggn - \nabla_\param^2 \log p(\param \given \model) $ \Comment{$\alpha_\model$ can model observation noise}
  \State $\model \gets \model + \gamma_\model \nabla_\model \sum_{n=1}^N \log p(\vy_n \given \hat{\vf}_n, \model) + \log p(\param \given \model) - \tfrac12 \log \left \vert \tfrac{1}{2\pi} \hat{\mH}_\param(\model) \right\vert$
  \State $\hat{\vh} \gets \textrm{vec}(\hat{\mH}\inv_\param)$ \Comment{Compute preconditioner of vjp; for \kfac see \cref{app:grad_kfac}}
  \State $\vg_{\augmentation} \gets \vzero$
  \For{each batch $\sB \subseteq \data$ of size $|\sB| \leq M$} \Comment{Aggregate invariance gradient}
    \State $\vg_{\augmentation} \gets \vg_{\augmentation} + \nabla_{\augmentation} \sum_{m \in \sB} \log p(\vy_m \given \vf(\vg(\vx_m,\vepsilon_s;\augmentation);\param))$
    \State $\hat{\vh}_{\sB}(\augmentation) \gets \textrm{vec}(\sum_{m \in \sB} \hat{\mH}_m^\ggn(\augmentation))$ \Comment{Compute with graph to \augmentation}
    \State $\vg_{\augmentation} \gets \vg_{\augmentation} + \textrm{vjp}(\hat{\vh}, \hat{\vh}_{\sB}(\augmentation), \augmentation)$ \Comment{vector-Jacobian product w.r.t~\augmentation}
  \EndFor
  \State $\augmentation \gets \augmentation + \gamma_{\augmentation} \vg_{\augmentation}$
\EndFor
\State \textbf{Return:} optimised neural network, invariance, and hyperparameters ($\param, \augmentation, \model$), approximation to log marginal likelihood $\log p(\data \given \augmentation, \model)$, and optionally posterior approximation $p(\param \given \data, \augmentation, \model)$. 
\end{algorithmic}
\end{algorithm}

%% file: sections/appendices/training_details.tex
The code for \ourmethod and the experiments is available at \url{https://github.com/tychovdo/lila}.

\subsection{Dataset Details}

We used MNIST \cite{lecun2010mnist}, FashionMNIST \cite{FashionMNIST} and CIFAR-10 \cite{krizhevsky2009learning} in our experiments. 
MNIST and FashionMNIST pixel values are scaled to the interval $[0,1]$ and CIFAR-10 images are standardised to zero mean and unit variance per channel following common practise.
We created the following transformed datasets to validate the invariance learning of our method:

\begin{itemize}
    \item Partially rotated dataset: Each sample rotated with randomly sampled radian angle from $U[-\frac{\pi}{2}, \frac{\pi}{2}]$.
    \item Fully rotated dataset: Each sample rotated with randomly sampled radian angle from $U[-\pi, \pi]$.
    \item Translated dataset: Each sample is translated by $dx$ pixels in x-direction and $dy$ pixels in the y-direction, both independently sampled from $dx, dy \sim U[-8, 8]$.
    \item Scaled dataset: Each sample scaled around center with $\exp(s)$ pixels in both x-direction and y-direction simultaneously, sampled from $s \sim U[-\log(2), \log(2)]$.
\end{itemize}

\subsection{Network Architectures}

For the MLP we use a single hidden layer with 1000 hidden units and a \texttt{tanh} activation function. For the CNN experiments we used a convolutional neural network with three convolutional layers with $3\times3$ filters, 1 stride, 1 padding, bias weights with increasing channels sizes (3-16-32-64) followed by a linear layer with 256 hidden units with \texttt{ReLU} activation function between layers.

For CIFAR-10, we used ResNets with fixup parameterization and initialization~\citep{zhang2019fixup} to avoid batch norm, which conflicts with a Bayesian model~\citep{wenzel2020good}.
The ResNets sizes are indicated by (input-channels - width) in the tables and figures.
We use ResNets (8-8) and (8-16) except in \cref{tab:large-resnet-results}, where we use a wide ResNet~\citep{zagoryuko2016wide} 16-4 with fixup~\citep{zhang2019fixup} as in \citep{daxberger2021laplace}.
The WRN 16-4 is only run on plain CIFAR-10 as they are expensive to run for all subset sizes and transformations.
All vanilla ResNets can fit the training data to $100\%$ accuracy.
Following \citet{immer2021scalable}, we use prior precision hyperparameters $\model$ per neural network layer and use a learning rate of $0.05$ for hyperparameters.

\subsection{Training Parameters}
For the MNIST and FashionMNIST experiments, we trained our models for 1000 epochs with a batch size of 1000 and 31 augmentation samples. The Adam optimizer was used with a learning rate of 0.005 cosine decayed to $0.0001$ and a momentum of $0.9$ for the network weights, and for the invariance and hyperparameters we used a learning rate of $0.05$ together with a 10 epochs burn-in period. Each experiment was repeated 3 times with different random seeds. For the MNIST and FashionMNIST subset of data results in \cref{app:additional-subset-experiments}, we use the same hyperparameters and 1000 epochs for subset sizes $[312, 1250, 5000, 20000]$.

For the CIFAR-10 experiments we trained the ResNets for $200$ epochs on the full data with initial learning rate of $0.1$ using SGD with momentum of $0.9$ and cosine learning rate decay to $10^{-6}$.
We use a batch size of $250$ and accumulate gradients with respect to augmentation parameters for $M=5000$ data points using $S=20$ augmentation samples.
We optimise $\augmentation$ using Adam~\citep{kingma2014adam} starting from epoch $10$ with learning rate $0.005$.
For Augerino~\citep{benton2020learning} we use the same learning rate, $0.005$ for \augmentation and default weight-decay of $10^{-4}$ as used in their experiments.
For the CIFAR-10 subset of data results in \cref{app:additional-subset-experiments}, we use the same hyperparameters but $[600, 400, 300, 250]$ epochs for subset sizes $[1000, 5000, 10000, 20000]$, respectively.
We report the mean and one standard error over three seeds for all experiments.

\subsection{Classification Example}
For the toy classification example with $N=200$ data points, we use a fully-connected neural network with a single hidden layer of $50$ neurons and \texttt{tanh} activation.
All methods train with $500$ steps with a learning rate of $0.1$ using the Adam optimiser~\citep{kingma2014adam}.
For the prior precision hyperparameter in $\model$, we use the same learning rate of $0.1$. 
For the augmentation parameter $\augmentation$, we use a lower learning rate of $0.005$ and decay it with cosine-decay to $0.0001$ due to the stochasticity of the gradients.
For data augmentation and our method, we use $S=100$ perturbation samples.
Each step, we optimise neural network parameters, invariance parameters $\augmentation$, and scale hyperparameters $\model$, which is a single prior precision in this case.
We plot the improved posterior predictive for Laplace approximations~\citep{immer2020improving} in both \cref{fig:illustration} and \cref{fig:toy_classification}.
The log marginal likelihood values given in \cref{fig:toy_classification} are all computed with a full Laplace-\ggn, even for the \kfac variant, for comparability of the estimates.
The hyperparameters $\model, \augmentation$ in \cref{fig:toy_classification}~(e) are optimised using \kfac for invariant neural networks proposed in \cref{sec:method}.

%% file: sections/appendices/toy_examples.tex
\begin{figure*}[!ht]
  \centering
  \includegraphics{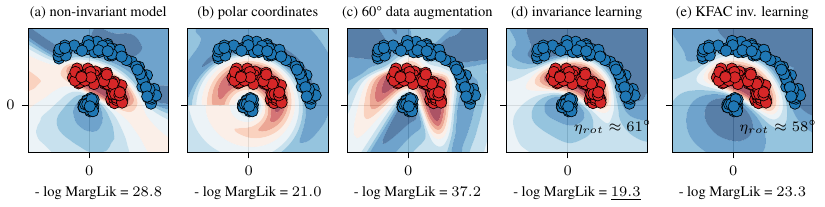}
  \caption{Different approaches to (approximately) invariant neural networks compared to a non-invariant one (a) and their corresponding Laplace log marginal likelihoods.
  (b): data representation in polar coordinates leads to a good marginal likelihood but requires prior knowledge.
  (c): standard data augmentation conflicts with the Bayesian model and achieves worse marginal likelihood.
  (d): treating the invariance or data augmentation parameter as part of a Bayesian model and optimising it allows to obtain the best marginal likelihood.
  (e): our efficient approximation based on an augmented \kfac Laplace approximation performs almost as well as the full Gauss-Newton in (d).}
  \label{fig:toy_classification}
\end{figure*}

In \cref{fig:toy_classification}, we compare our approach using the standard \ggn and the \kfac-\ggn with three baselines on a classification example that was generated with a soft $\pm 60^\circ$ rotational invariance around the origin.
We consider only rotational invariance about the origin with parameter $\eta_\textrm{rot}$.
Our method with both full and \kfac \ggn successfully learns the rotational invariance and obtains a similar value for $\eta_\textrm{rot}$ as the data generating process and obtains the best marginal likelihood indicating a better generalisation.
The non-invariant model achieves a worse marginal likelihood and standard data augmentation leads to an even lower value due to problems with its likelihood~\citep{nabarro2021data}.
A model in polar coordinates corresponds to incorporating prior knowledge manually and, as expected, performs on par with our learned model.
Our model, which can explicitly learn invariances, attains a slightly better log marginal likelihood since the polar model is not restricted to be fully invariant as the prediction can vary along the rotational angle.

%% file: sections/appendices/quantitative_results.tex
\input{sections/table_full}

%% file: sections/table_full.tex
\begin{table*}[!h]
    \centering
    \caption{Marginal Likelihood and Test Accuracy for models using GGN-KFAC on different versions of the MNIST, FashionMNIST and CIFAR-10 datasets. Our proposed method outperforms the non-invariants for almost all models and datasets, in terms of both measures.}
    \begin{adjustbox}{max width=1.0\linewidth}
    \begin{tabular}{l l l|c c c c c|c c c c c}
    &
    &
       & \multicolumn{5}{ c }{\underline{Marginal Likelihood}} 
       & \multicolumn{5}{ c }{\underline{Test Accuracy}}\\
      Network
      & Dataset
      & Model
      & \shortstack{{\footnotesize Fully rotated} \\ Dataset}
      & \shortstack{{\footnotesize Partially rotated} \\ Dataset}
      & \shortstack{Translated \\ Dataset}
      & \shortstack{Scaled \\ Dataset}
      & \shortstack{Original \\ Dataset}
      & \shortstack{{\footnotesize Fully rotated} \\ Dataset}
      & \shortstack{{\footnotesize Partially rotated} \\ Dataset}
      & \shortstack{Translated \\ Dataset}
      & \shortstack{Scaled \\ Dataset}
      & \shortstack{Original \\ Dataset}
      \\
    \hline
    \multirow{6}{*}{MNIST} & 
    \multirow{3}{*}{MLP} & 
    \multicolumn{1}{|l|}{non-invariant}
  & -31.3k \pms{99} & -23.1k \pms{29} & -36.0k \pms{10} & -17.5k \pms{28} & -10.5k \pms{5}
 & 93.82 \pms{0.10} & 95.83 \pms{0.03} & 94.15 \pms{0.02} & 97.07 \pms{0.06} & 98.20 \pms{0.03}
      \\
    &
    & 
    \multicolumn{1}{|l|}{Augerino}
 & - & - & - & - & - 
 & \textbf{97.83 \pms{0.03}} & 96.35 \pms{0.02} & 94.47 \pms{0.08} & 97.45 \pms{0.03} & 98.45 \pms{0.03}
      \\
    &
    & 
    \multicolumn{1}{|l|}{\ourmethod (\textbf{ours})}
  & \textbf{-9.7k \pms{25}} & \textbf{-10.5k \pms{27}} & \textbf{-14.5k \pms{11}} & \textbf{-11.3k \pms{22}} & \textbf{-6.2k \pms{33}}
 & 97.74 \pms{0.07} & \textbf{97.81 \pms{0.11}} & \textbf{97.28 \pms{0.05}} & \textbf{98.33 \pms{0.05}} & \textbf{98.98 \pms{0.05}}
      \\
    \cline{2-13}
    &
    \multirow{3}{*}{CNN} & 
    \multicolumn{1}{|l|}{non-invariant}
     & -14.6k \pms{1626} & -10.7k \pms{1070} & -16.5k \pms{1014} & -9.5k \pms{225} & -5.0k \pms{428}
 & 95.97 \pms{0.33} & 97.51 \pms{0.17} & 96.54 \pms{0.29} & 98.37 \pms{0.00} & 99.09 \pms{0.02}
      \\
    & 
    &
    \multicolumn{1}{|l|}{Augerino}
 & - & - & - & - & - 
 & \textbf{99.04 \pms{0.02}} & 98.91 \pms{0.03} & 97.79 \pms{0.09} & 98.77 \pms{0.06} & 98.26 \pms{0.10}
      \\
    &
    & 
    \multicolumn{1}{|l|}{\ourmethod (\textbf{ours})}
     & \textbf{-6.9k \pms{77}} & \textbf{-7.2k \pms{295}} & \textbf{-8.6k \pms{136}} & \textbf{-7.5k \pms{54}} & \textbf{-4.3k \pms{27}}
 & 98.83 \pms{0.07} & \textbf{98.92 \pms{0.05}} & \textbf{98.69 \pms{0.07}} & \textbf{99.01 \pms{0.06}} & \textbf{99.42 \pms{0.02}}
      \\
    \hline
    \multirow{6}{*}{F-MNIST} & 
    \multirow{3}{*}{MLP} & 
    \multicolumn{1}{|l|}{non-invariant}
     & -55.1k \pms{117} & -46.2k \pms{184} & -52.4k \pms{36} & -39.2k \pms{27} & -25.5k \pms{6}
 & 77.62 \pms{0.30} & 81.10 \pms{0.23} & 77.68 \pms{0.10} & 81.84 \pms{0.05} & 88.48 \pms{0.56}
      \\
    &
    & 
    \multicolumn{1}{|l|}{Augerino}
 & - & - & - & - & - 
 & 77.76 \pms{0.15} & 81.40 \pms{0.05} & 78.05 \pms{0.10} & 82.46 \pms{0.09} & 89.10 \pms{0.13}
      \\
    &
    & 
    \multicolumn{1}{|l|}{\ourmethod (\textbf{ours})}
  & \textbf{-29.3k \pms{63}} & \textbf{-29.4k \pms{148}} & \textbf{-36.6k \pms{17}} & \textbf{-34.7k \pms{201}} & \textbf{-22.6k \pms{42}}
 & \textbf{87.39 \pms{0.06}} & \textbf{86.72 \pms{0.13}} & \textbf{84.62 \pms{0.08}} & \textbf{84.31 \pms{0.06}} & \textbf{89.94 \pms{0.12}}
      \\
    \cline{2-13}
    &
    \multirow{3}{*}{CNN} & 
    \multicolumn{1}{|l|}{non-invariant}
     & -58.7k \pms{189} & -49.9k \pms{275} & -53.8k \pms{257} & -43.3k \pms{488} & -28.9k \pms{165}
 & 78.69 \pms{0.28} & 82.12 \pms{0.35} & 80.33 \pms{0.19} & 83.66 \pms{0.37} & 89.54 \pms{0.23}
      \\
    & 
    &
    \multicolumn{1}{|l|}{Augerino}
 & - & - & - & - & - 
 & 85.76 \pms{3.23} & 81.54 \pms{0.19} & 82.94 \pms{0.13} & 83.58 \pms{0.08} & 90.07 \pms{0.12}
      \\
    &
    & 
    \multicolumn{1}{|l|}{\ourmethod (\textbf{ours})}
  & \textbf{-26.5k \pms{101}} & \textbf{-27.2k \pms{0}} & \textbf{-29.2k \pms{65}} & \textbf{-30.1k \pms{0}} & \textbf{-21.0k \pms{233}}
 & \textbf{89.45 \pms{0.03}} & \textbf{88.40 \pms{0.00}} & \textbf{87.73 \pms{0.20}} & \textbf{87.33 \pms{0.00}} & \textbf{91.92 \pms{0.21}}
      \\
    \hline
   \multirow{6}{*}{CIFAR-10} & 
    \multirow{3}{*}{ResNet (8-8)} & 
    \multicolumn{1}{|l|}{non-invariant}
     & -73.1k \pms{152} & -67.5k \pms{334} & -53.7k \pms{154} & -59.3k \pms{749} & -46.5k \pms{345}
 & 51.14 \pms{0.47} & 55.29 \pms{0.70} & 64.84 \pms{0.21} & 59.81 \pms{0.96} & 69.74 \pms{0.75}
      \\
    & 
    &
    \multicolumn{1}{|l|}{Augerino}
    & - & - & - & - & - 
 & 70.88 \pms{0.17} & 70.95 \pms{0.26} & \textbf{74.44 \pms{0.24}} & 70.67 \pms{0.28} & 79.34 \pms{0.36}
      \\
    &
    & 
    \multicolumn{1}{|l|}{\ourmethod (\textbf{ours})}
     & \textbf{-51.1k \pms{244}} & \textbf{-47.8k \pms{296}} & \textbf{-38.6k \pms{181}} & \textbf{-41.7k \pms{225}} & \textbf{-31.2k \pms{346}}
 & \textbf{71.06 \pms{0.47}} & \textbf{73.03 \pms{0.45}} & 74.18 \pms{0.07} & \textbf{71.54 \pms{0.18}} & \textbf{80.22 \pms{0.24}}
      \\
    \cline{2-13}
    &
    \multirow{3}{*}{ResNet (8-16)} & 
    \multicolumn{1}{|l|}{non-invariant}
     & -80.9k \pms{483} & -72.8k \pms{133} & -56.2k \pms{329} & -62.0k \pms{283} & -46.1k \pms{953}
 & 54.16 \pms{0.40} & 59.90 \pms{0.12} & 69.65 \pms{0.16} & 66.06 \pms{0.13} & 74.13 \pms{0.51}
      \\
    & 
    &
    \multicolumn{1}{|l|}{Augerino}
    & - & - & - & - & - 
 & 75.40 \pms{0.19} & 74.76 \pms{0.34} & 73.71 \pms{0.31} & 72.07 \pms{0.09} & 79.03 \pms{1.04}
      \\
    &
    & 
    \multicolumn{1}{|l|}{\ourmethod (\textbf{ours})}
     & \textbf{-43.1k \pms{217}} & \textbf{-38.7k \pms{135}} & \textbf{-35.4k \pms{1969}} & \textbf{-40.2k \pms{3404}} & \textbf{-30.2k \pms{2344}}
 & \textbf{79.50 \pms{0.62}} & \textbf{77.71 \pms{0.46}} & \textbf{79.21 \pms{0.17}} & \textbf{76.03 \pms{0.15}} & \textbf{84.19 \pms{0.76}}
      \\
    \hline
    \end{tabular}
    \end{adjustbox}
    \label{tab:quantitative-table-full}
\end{table*}

%% file: sections/appendices/additional_trajectories.tex
\begin{figure*}[h!]
    \centering
    \includegraphics[width=0.85\textwidth]{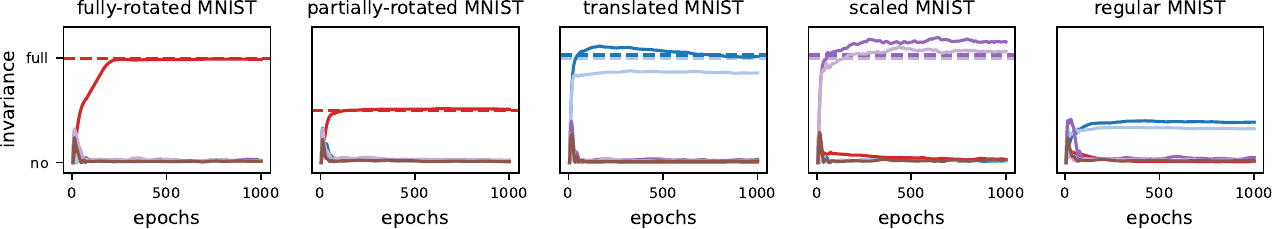}
    \caption{MLP on MNIST}
\end{figure*}
\begin{figure*}[h!]
    \centering
    \includegraphics[width=0.85\textwidth]{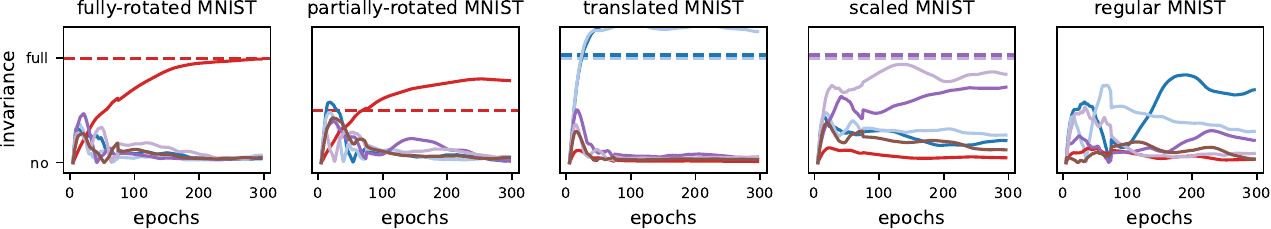}
    \caption{CNN on MNIST}
\end{figure*}
\begin{figure*}[h!]
    \centering
    \includegraphics[width=0.85\textwidth]{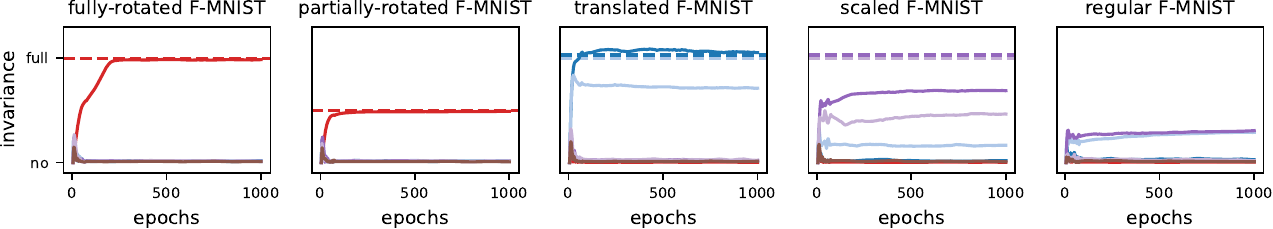}
    \caption{MLP on F-MNIST}
\end{figure*}
\begin{figure*}[h!]
    \centering
    \includegraphics[width=0.85\textwidth]{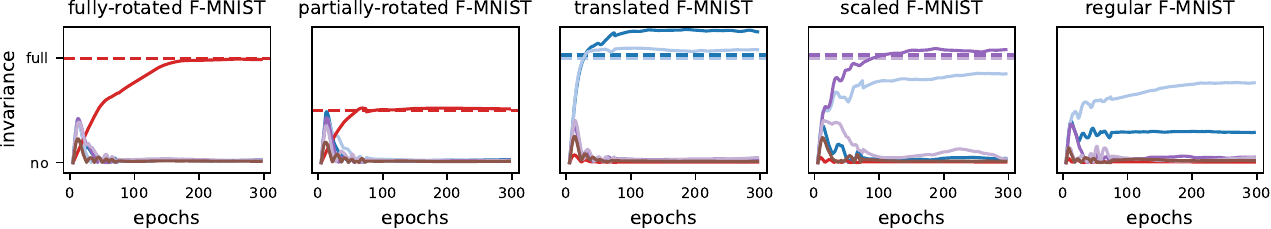}
    \caption{CNN on F-MNIST}
\end{figure*}
\begin{figure*}[h!]
    \centering
    \includegraphics[width=0.85\textwidth]{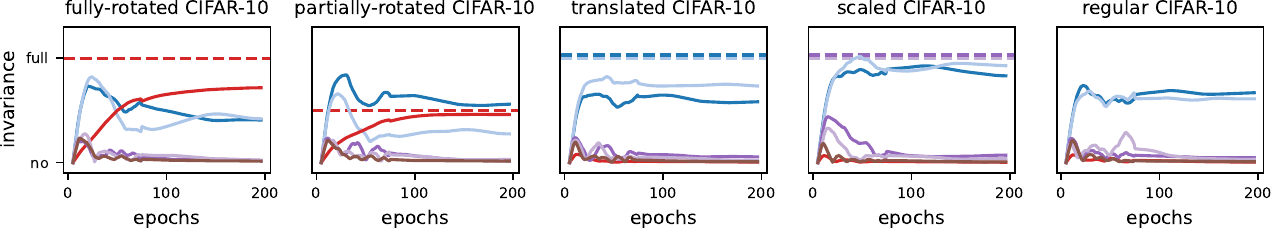}
    \caption{ResNet on CIFAR-10}
\end{figure*}

%% file: sections/appendices/additional_barplots.tex
\begin{figure*}[!h]
    \centering
    \includegraphics[width=0.61\textwidth]{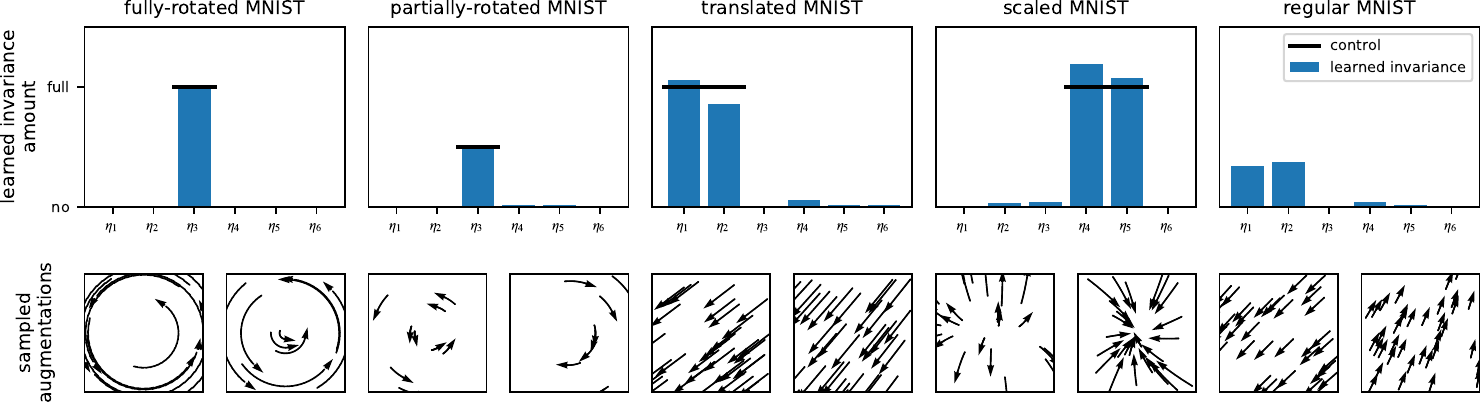}
    \caption{MLP on MNIST}
\end{figure*}

\begin{figure*}[!h]
    \centering
    \includegraphics[width=0.61\textwidth]{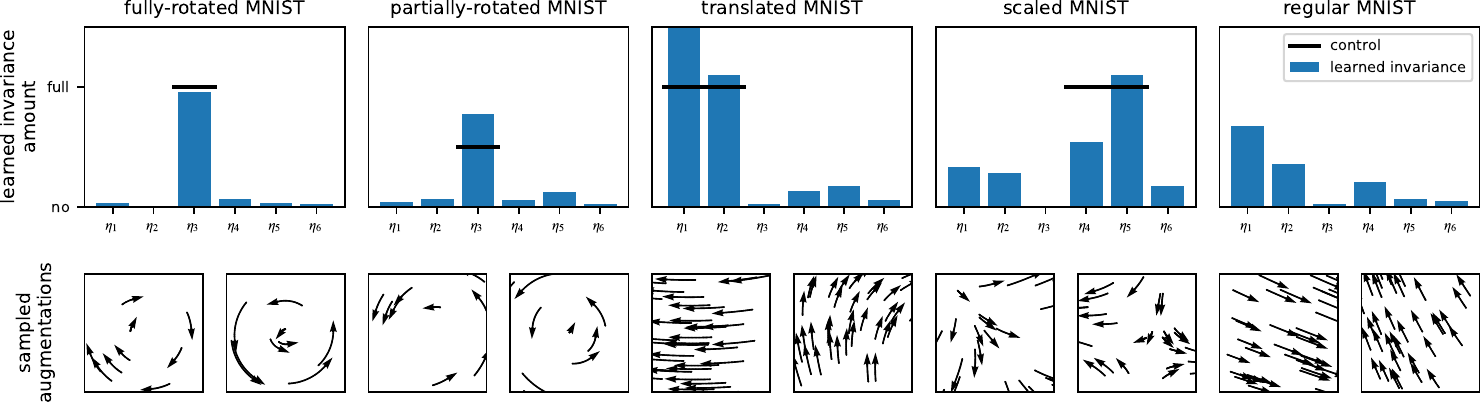}
    \caption{CNN on MNIST}
\end{figure*}

\begin{figure*}[!h]
    \centering
    \includegraphics[width=0.61\textwidth]{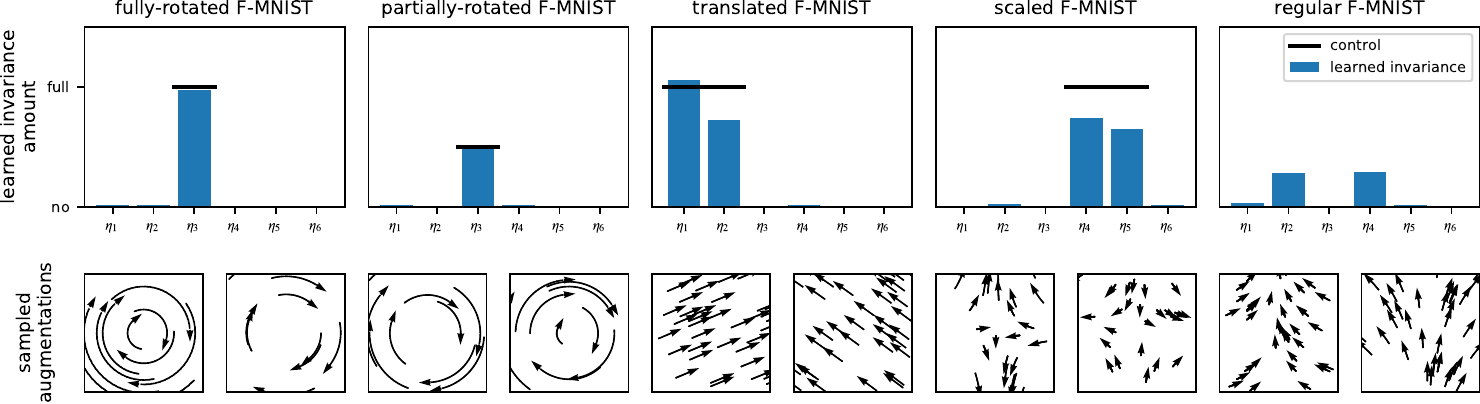}
    \caption{MLP on F-MNIST}
\end{figure*}

\begin{figure*}[!h]
    \centering
    \includegraphics[width=0.61\textwidth]{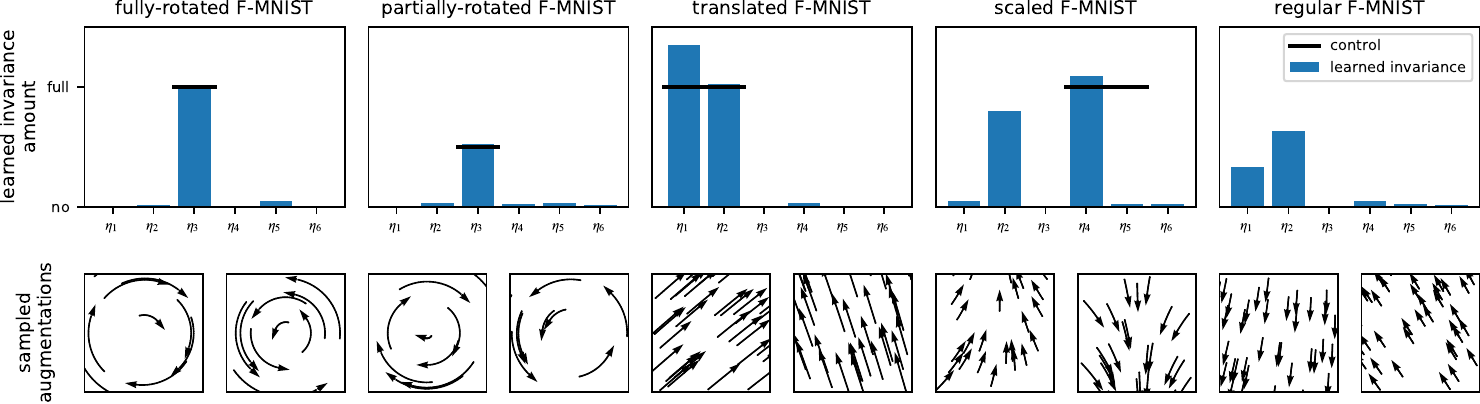}
    \caption{CNN on F-MNIST}
\end{figure*}

\begin{figure*}[!h]
    \centering
    \includegraphics[width=0.61\textwidth]{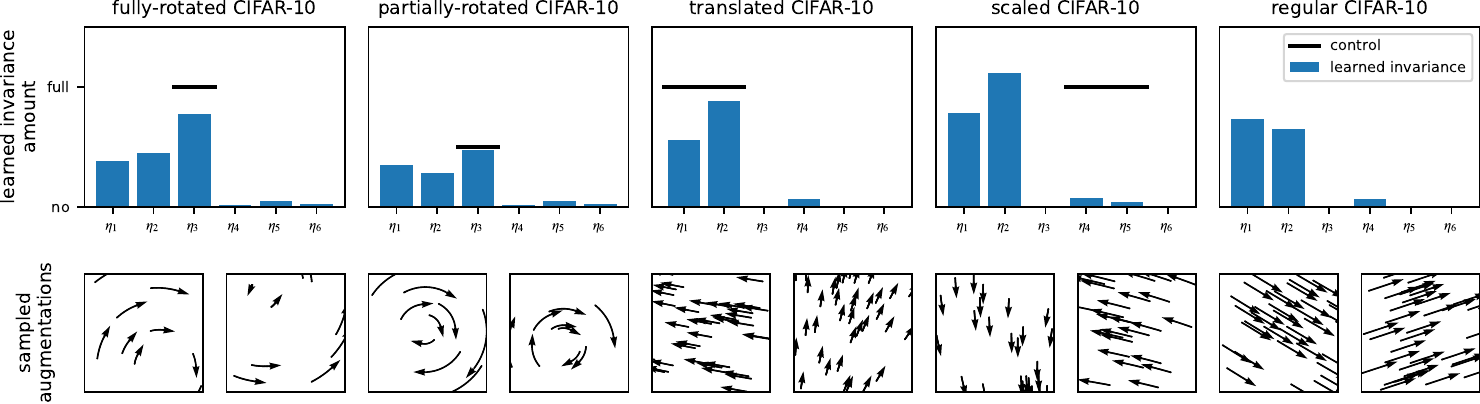}
    \caption{ResNet on CIFAR-10}
\end{figure*}

%% file: sections/appendices/additional_subsets.tex
\begin{figure*}[!h]
    \centering
    \includegraphics[width=0.9\textwidth]{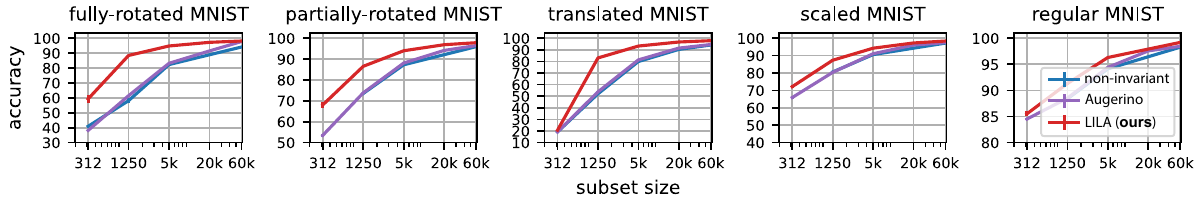}
    \caption{MLP on MNIST}
\end{figure*}

\begin{figure*}[!h]
    \centering
    \includegraphics[width=0.9\textwidth]{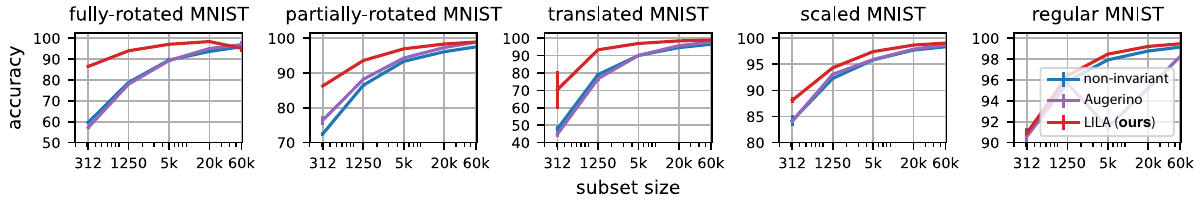}
    \caption{CNN on MNIST}
\end{figure*}

\begin{figure*}[!h]
    \centering
    \includegraphics[width=0.9\textwidth]{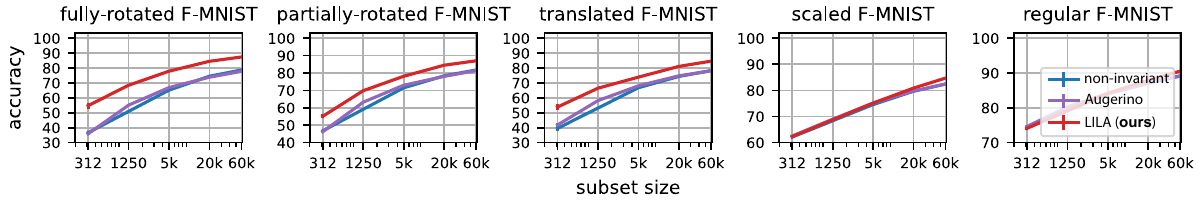}
    \caption{MLP on F-MNIST}
\end{figure*}

\begin{figure*}[!h]
    \centering
    \includegraphics[width=0.9\textwidth]{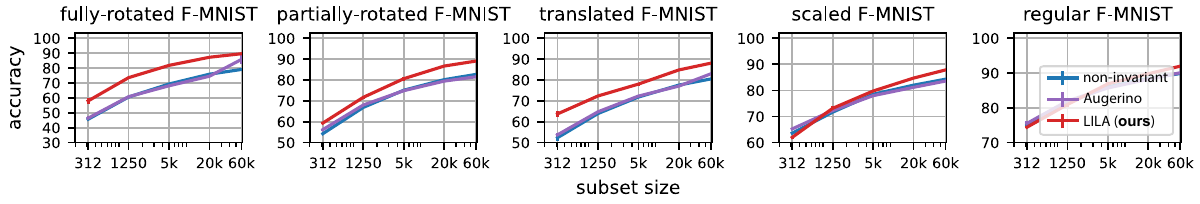}
    \caption{CNN on F-MNIST}
\end{figure*}

%% file: main_neurips2022.bbl
\begin{thebibliography}{64}
\providecommand{\natexlab}[1]{#1}
\providecommand{\url}[1]{\texttt{#1}}
\expandafter\ifx\csname urlstyle\endcsname\relax
  \providecommand{\doi}[1]{doi: #1}\else
  \providecommand{\doi}{doi: \begingroup \urlstyle{rm}\Url}\fi

\bibitem[Antor{\'a}n et~al.(2022{\natexlab{a}})Antor{\'a}n, Barbano, Leuschner,
  Hern{\'a}ndez-Lobato, and Jin]{antoran2022probabilistic}
Javier Antor{\'a}n, Riccardo Barbano, Johannes Leuschner, Jos{\'e}~Miguel
  Hern{\'a}ndez-Lobato, and Bangti Jin.
\newblock A probabilistic deep image prior for computational tomography.
\newblock \emph{arXiv preprint arXiv:2203.00479}, 2022{\natexlab{a}}.

\bibitem[Antor{\'a}n et~al.(2022{\natexlab{b}})Antor{\'a}n, Janz, Allingham,
  Daxberger, Barbano, Nalisnick, and Hern{\'a}ndez-Lobato]{antoran2022adapting}
Javier Antor{\'a}n, David Janz, James~U Allingham, Erik Daxberger, Riccardo~Rb
  Barbano, Eric Nalisnick, and Jos{\'e}~Miguel Hern{\'a}ndez-Lobato.
\newblock Adapting the linearised laplace model evidence for modern deep
  learning.
\newblock In \emph{International Conference on Machine Learning},
  2022{\natexlab{b}}.

\bibitem[Batzner et~al.(2021)Batzner, Musaelian, Sun, Geiger, Mailoa,
  Kornbluth, Molinari, Smidt, and Kozinsky]{batzner2021se}
Simon Batzner, Albert Musaelian, Lixin Sun, Mario Geiger, Jonathan~P Mailoa,
  Mordechai Kornbluth, Nicola Molinari, Tess~E Smidt, and Boris Kozinsky.
\newblock Se (3)-equivariant graph neural networks for data-efficient and
  accurate interatomic potentials.
\newblock \emph{arXiv preprint arXiv:2101.03164}, 2021.

\bibitem[Benton et~al.(2020)Benton, Finzi, Izmailov, and
  Wilson]{benton2020learning}
Gregory Benton, Marc Finzi, Pavel Izmailov, and Andrew~Gordon Wilson.
\newblock Learning invariances in neural networks.
\newblock \emph{arXiv preprint arXiv:2010.11882}, 2020.

\bibitem[Bishop(2006)]{bishop2006pattern}
Christopher~M Bishop.
\newblock \emph{Pattern recognition and machine learning}.
\newblock Information Science and Statistics. Springer, 2006.

\bibitem[Blundell et~al.(2015)Blundell, Cornebise, Kavukcuoglu, and
  Wierstra]{blundell2015weight}
Charles Blundell, Julien Cornebise, Koray Kavukcuoglu, and Daan Wierstra.
\newblock Weight uncertainty in neural networks.
\newblock In \emph{Proceedings of the 32nd International Conference on Machine
  Learning}, pages 1613--1622, 2015.

\bibitem[Botev et~al.(2017)Botev, Ritter, and Barber]{botev2017practical}
Aleksandar Botev, Hippolyt Ritter, and David Barber.
\newblock Practical {G}auss-{N}ewton optimisation for deep learning.
\newblock In \emph{International Conference on Machine Learning}, International
  Convention Centre, Sydney, Australia, 2017. PMLR.

\bibitem[Bottou(2010)]{bottou2010large}
L{\'e}on Bottou.
\newblock Large-scale machine learning with stochastic gradient descent.
\newblock In \emph{Proceedings of COMPSTAT'2010}, pages 177--186. Springer,
  2010.

\bibitem[Brandstetter et~al.(2021)Brandstetter, Hesselink, van~der Pol,
  Bekkers, and Welling]{brandstetter2021geometric}
Johannes Brandstetter, Rob Hesselink, Elise van~der Pol, Erik Bekkers, and Max
  Welling.
\newblock Geometric and physical quantities improve e (3) equivariant message
  passing.
\newblock \emph{arXiv preprint arXiv:2110.02905}, 2021.

\bibitem[Cohen and Welling(2016)]{cohen2016group}
Taco Cohen and Max Welling.
\newblock Group equivariant convolutional networks.
\newblock In \emph{International conference on machine learning}, pages
  2990--2999. PMLR, 2016.

\bibitem[Cohen et~al.(2018)Cohen, Geiger, K{\"o}hler, and
  Welling]{cohen2018spherical}
Taco~S Cohen, Mario Geiger, Jonas K{\"o}hler, and Max Welling.
\newblock Spherical cnns.
\newblock \emph{arXiv preprint arXiv:1801.10130}, 2018.

\bibitem[Cubuk et~al.(2018)Cubuk, Zoph, Mane, Vasudevan, and
  Le]{cubuk2018autoaugment}
Ekin~D Cubuk, Barret Zoph, Dandelion Mane, Vijay Vasudevan, and Quoc~V Le.
\newblock Autoaugment: Learning augmentation policies from data.
\newblock \emph{arXiv preprint arXiv:1805.09501}, 2018.

\bibitem[Dangel et~al.(2019)Dangel, Kunstner, and Hennig]{dangel2019backpack}
Felix Dangel, Frederik Kunstner, and Philipp Hennig.
\newblock Backpack: Packing more into backprop.
\newblock In \emph{Proceedings of 7th International Conference on Learning
  Representations}, 2019.

\bibitem[Daxberger et~al.(2021)Daxberger, Kristiadi, Immer, Eschenhagen, Bauer,
  and Hennig]{daxberger2021laplace}
Erik Daxberger, Agustinus Kristiadi, Alexander Immer, Runa Eschenhagen,
  Matthias Bauer, and Philipp Hennig.
\newblock Laplace redux-effortless bayesian deep learning.
\newblock \emph{Advances in Neural Information Processing Systems}, 34, 2021.

\bibitem[Dehaene(2019)]{dehaene2019deterministic}
Guillaume~P Dehaene.
\newblock A deterministic and computable bernstein-von mises theorem.
\newblock \emph{arXiv preprint arXiv:1904.02505}, 2019.

\bibitem[Fong and Holmes(2020)]{fong2020marginal}
Edwin Fong and CC~Holmes.
\newblock On the marginal likelihood and cross-validation.
\newblock \emph{Biometrika}, 107\penalty0 (2):\penalty0 489--496, 2020.

\bibitem[Foong et~al.(2019)Foong, Li, Hern{\'a}ndez-Lobato, and
  Turner]{foong2019between}
Andrew~YK Foong, Yingzhen Li, Jos{\'e}~Miguel Hern{\'a}ndez-Lobato, and
  Richard~E Turner.
\newblock 'in-between'uncertainty in bayesian neural networks.
\newblock \emph{arXiv preprint arXiv:1906.11537}, 2019.

\bibitem[Fukushima and Miyake(1982)]{fukushima1982neocognitron}
Kunihiko Fukushima and Sei Miyake.
\newblock Neocognitron: A self-organizing neural network model for a mechanism
  of visual pattern recognition.
\newblock In \emph{Competition and cooperation in neural nets}, pages 267--285.
  Springer, 1982.

\bibitem[Germain et~al.(2016)Germain, Bach, Lacoste, and
  Lacoste-Julien]{germain2016pac}
Pascal Germain, Francis Bach, Alexandre Lacoste, and Simon Lacoste-Julien.
\newblock Pac-bayesian theory meets bayesian inference.
\newblock In D.~Lee, M.~Sugiyama, U.~Luxburg, I.~Guyon, and R.~Garnett,
  editors, \emph{Advances in Neural Information Processing Systems}, volume~29.
  Curran Associates, Inc., 2016.

\bibitem[Ginsbourger et~al.(2016)Ginsbourger, Roustant, and
  Durrande]{ginsbourger2016}
David Ginsbourger, Olivier Roustant, and Nicolas Durrande.
\newblock On degeneracy and invariances of random fields paths with
  applications in gaussian process modelling.
\newblock \emph{Journal of Statistical Planning and Inference}, 170:\penalty0
  117--128, 2016.
\newblock ISSN 0378-3758.

\bibitem[Gr{\"u}nwald(2007)]{grunwald2007minimum}
Peter~D Gr{\"u}nwald.
\newblock \emph{The minimum description length principle}.
\newblock MIT press, 2007.

\bibitem[He et~al.(2016)He, Zhang, Ren, and Sun]{he2016deep}
Kaiming He, Xiangyu Zhang, Shaoqing Ren, and Jian Sun.
\newblock Deep residual learning for image recognition.
\newblock In \emph{Proceedings of the IEEE conference on computer vision and
  pattern recognition}, pages 770--778, 2016.

\bibitem[Immer et~al.(2021{\natexlab{a}})Immer, Bauer, Fortuin, R{\"a}tsch, and
  Khan]{immer2021scalable}
Alexander Immer, Matthias Bauer, Vincent Fortuin, Gunnar R{\"a}tsch, and
  Mohammad~Emtiyaz Khan.
\newblock Scalable marginal likelihood estimation for model selection in deep
  learning.
\newblock \emph{arXiv preprint arXiv:2104.04975}, 2021{\natexlab{a}}.

\bibitem[Immer et~al.(2021{\natexlab{b}})Immer, Korzepa, and
  Bauer]{immer2020improving}
Alexander Immer, Maciej Korzepa, and Matthias Bauer.
\newblock Improving predictions of bayesian neural nets via local
  linearization.
\newblock In \emph{Proceedings of The 24th International Conference on
  Artificial Intelligence and Statistics}, pages 703--711, 2021{\natexlab{b}}.

\bibitem[Immer et~al.(2022)Immer, Torroba~Hennigen, Fortuin, and
  Cotterell]{immer2022probing}
Alexander Immer, Lucas Torroba~Hennigen, Vincent Fortuin, and Ryan Cotterell.
\newblock Probing as quantifying inductive bias.
\newblock In \emph{Proceedings of the 60th Annual Meeting of the Association
  for Computational Linguistics}, pages 1839--1851, 2022.

\bibitem[Jacot et~al.(2018)Jacot, Gabriel, and Hongler]{jacot2018neural}
Arthur Jacot, Franck Gabriel, and Cl{\'e}ment Hongler.
\newblock Neural tangent kernel: Convergence and generalization in neural
  networks.
\newblock In \emph{Advances in neural information processing systems}, pages
  8571--8580, 2018.

\bibitem[Jaderberg et~al.(2015)Jaderberg, Simonyan, Zisserman,
  et~al.]{jaderberg2015spatial}
Max Jaderberg, Karen Simonyan, Andrew Zisserman, et~al.
\newblock Spatial transformer networks.
\newblock \emph{Advances in neural information processing systems},
  28:\penalty0 2017--2025, 2015.

\bibitem[Khan et~al.(2019)Khan, Immer, Abedi, and Korzepa]{khan2019approximate}
Mohammad Emtiyaz~E Khan, Alexander Immer, Ehsan Abedi, and Maciej Korzepa.
\newblock Approximate inference turns deep networks into gaussian processes.
\newblock In \emph{Advances in Neural Information Processing Systems}, pages
  3088--3098, 2019.

\bibitem[Kingma and Ba(2015)]{kingma2014adam}
Diederik~P Kingma and Jimmy Ba.
\newblock Adam: A method for stochastic optimization.
\newblock In \emph{International Conference on Learning Representations}, 2015.

\bibitem[Kingma and Welling(2013)]{kingma2013auto}
Diederik~P Kingma and Max Welling.
\newblock Auto-encoding variational bayes.
\newblock \emph{arXiv preprint arXiv:1312.6114}, 2013.

\bibitem[Kondor(2008)]{kondorthesis}
Imre~Risi Kondor.
\newblock Group theoretical methods in machine learning, 2008.

\bibitem[Kondor et~al.(2018)Kondor, Lin, and Trivedi]{kondor2018clebsch}
Risi Kondor, Zhen Lin, and Shubhendu Trivedi.
\newblock Clebsch--gordan nets: a fully fourier space spherical convolutional
  neural network.
\newblock \emph{Advances in Neural Information Processing Systems},
  31:\penalty0 10117--10126, 2018.

\bibitem[Krizhevsky et~al.(2009)Krizhevsky, Hinton,
  et~al.]{krizhevsky2009learning}
Alex Krizhevsky, Geoffrey Hinton, et~al.
\newblock Learning multiple layers of features from tiny images.
\newblock 2009.

\bibitem[Kunstner et~al.(2019)Kunstner, Hennig, and
  Balles]{kunstner2019limitations}
Frederik Kunstner, Philipp Hennig, and Lukas Balles.
\newblock Limitations of the empirical fisher approximation for natural
  gradient descent.
\newblock In \emph{Advances in Neural Information Processing Systems}, pages
  4158--4169, 2019.

\bibitem[Laplace(1774)]{laplace1774memoire}
Pierre-Simon~de Laplace.
\newblock M{\'e}moire sur la probabilit{\'e} des causes par les
  {\'e}v{\'e}nements.
\newblock \emph{Mémoires de l’Académie royale des sciences de Paris
  (Savants étrangers)}, 6:\penalty0 621--656, 1774.

\bibitem[LeCun and Cortes(2010)]{lecun2010mnist}
Yann LeCun and Corinna Cortes.
\newblock {MNIST} handwritten digit database.
\newblock http://yann.lecun.com/exdb/mnist/, 2010.
\newblock URL \url{http://yann.lecun.com/exdb/mnist/}.

\bibitem[LeCun et~al.(1998)LeCun, Bottou, Bengio, and
  Haffner]{lecun1998gradient}
Yann LeCun, L{\'e}on Bottou, Yoshua Bengio, and Patrick Haffner.
\newblock Gradient-based learning applied to document recognition.
\newblock \emph{Proceedings of the IEEE}, 86\penalty0 (11):\penalty0
  2278--2324, 1998.

\bibitem[Lorraine et~al.(2020)Lorraine, Vicol, and Duvenaud]{lorraine2020}
Jonathan Lorraine, Paul Vicol, and David Duvenaud.
\newblock Optimizing millions of hyperparameters by implicit differentiation.
\newblock In Silvia Chiappa and Roberto Calandra, editors, \emph{Proceedings of
  the Twenty Third International Conference on Artificial Intelligence and
  Statistics}, volume 108 of \emph{Proceedings of Machine Learning Research},
  pages 1540--1552. PMLR, 26--28 Aug 2020.

\bibitem[MacKay(1992)]{mackay1992practical}
David~JC MacKay.
\newblock A practical bayesian framework for backpropagation networks.
\newblock \emph{Neural computation}, 4\penalty0 (3):\penalty0 448--472, 1992.

\bibitem[MacKay(2003)]{mackay2003information}
David~JC MacKay.
\newblock \emph{Information theory, inference and learning algorithms}.
\newblock Cambridge university press, 2003.

\bibitem[Martens(2020)]{martens2014new}
James Martens.
\newblock New insights and perspectives on the natural gradient method.
\newblock \emph{Journal of Machine Learning Research}, 21\penalty0
  (146):\penalty0 1--76, 2020.

\bibitem[Martens and Grosse(2015)]{martens2015optimizing}
James Martens and Roger Grosse.
\newblock Optimizing neural networks with kronecker-factored approximate
  curvature.
\newblock In \emph{International conference on machine learning}, pages
  2408--2417, 2015.

\bibitem[Moler and Van~Loan(2003)]{moler2003nineteen}
Cleve Moler and Charles Van~Loan.
\newblock Nineteen dubious ways to compute the exponential of a matrix,
  twenty-five years later.
\newblock \emph{SIAM review}, 45\penalty0 (1):\penalty0 3--49, 2003.

\bibitem[Murphy(2012)]{murphy2012machine}
Kevin~P Murphy.
\newblock \emph{Machine learning: a probabilistic perspective}.
\newblock MIT press, 2012.

\bibitem[Nabarro et~al.(2021)Nabarro, Ganev, Garriga-Alonso, Fortuin, van~der
  Wilk, and Aitchison]{nabarro2021data}
Seth Nabarro, Stoil Ganev, Adri{\`a} Garriga-Alonso, Vincent Fortuin, Mark
  van~der Wilk, and Laurence Aitchison.
\newblock Data augmentation in bayesian neural networks and the cold posterior
  effect.
\newblock \emph{arXiv preprint arXiv:2106.05586}, 2021.

\bibitem[Ober and Aitchison(2021)]{ober2021global}
Sebastian~W Ober and Laurence Aitchison.
\newblock Global inducing point variational posteriors for bayesian neural
  networks and deep gaussian processes.
\newblock In \emph{International Conference on Machine Learning}, pages
  8248--8259. PMLR, 2021.

\bibitem[Ober et~al.(2021)Ober, Rasmussen, and van~der Wilk]{ober2021promises}
Sebastian~W. Ober, Carl~E. Rasmussen, and Mark van~der Wilk.
\newblock The promises and pitfalls of deep kernel learning.
\newblock In Cassio de~Campos and Marloes~H. Maathuis, editors,
  \emph{Proceedings of the Thirty-Seventh Conference on Uncertainty in
  Artificial Intelligence (UAI)}, volume 161 of \emph{Proceedings of Machine
  Learning Research}, pages 1206--1216. PMLR, 27--30 Jul 2021.

\bibitem[Osawa et~al.(2019)Osawa, Swaroop, Khan, Jain, Eschenhagen, Turner, and
  Yokota]{osawa2019practical}
Kazuki Osawa, Siddharth Swaroop, Mohammad Emtiyaz~E Khan, Anirudh Jain, Runa
  Eschenhagen, Richard~E Turner, and Rio Yokota.
\newblock Practical deep learning with bayesian principles.
\newblock In \emph{Advances in Neural Information Processing Systems}, pages
  4289--4301, 2019.

\bibitem[Osawa et~al.(2020)Osawa, Tsuji, Ueno, Naruse, Foo, and
  Yokota]{osawa2020scalable}
Kazuki Osawa, Yohei Tsuji, Yuichiro Ueno, Akira Naruse, Chuan-Sheng Foo, and
  Rio Yokota.
\newblock Scalable and practical natural gradient for large-scale deep
  learning.
\newblock \emph{IEEE Transactions on Pattern Analysis and Machine
  Intelligence}, 2020.

\bibitem[Paszke et~al.(2017)Paszke, Gross, Chintala, Chanan, Yang, DeVito, Lin,
  Desmaison, Antiga, and Lerer]{paszke2017automatic}
Adam Paszke, Sam Gross, Soumith Chintala, Gregory Chanan, Edward Yang, Zachary
  DeVito, Zeming Lin, Alban Desmaison, Luca Antiga, and Adam Lerer.
\newblock Automatic differentiation in pytorch.
\newblock 2017.

\bibitem[Rasmussen and Ghahramani(2001)]{rasmussen2001occam}
Carl~Edward Rasmussen and Zoubin Ghahramani.
\newblock Occam's razor.
\newblock In \emph{Advances in neural information processing systems}, pages
  294--300, 2001.

\bibitem[Rasmussen and Williams(2006)]{rasmussen2006gaussian}
Carl~Edward Rasmussen and Christopher~KI Williams.
\newblock \emph{Gaussian processes for machine learning}.
\newblock MIT press Cambridge, MA, 2006.

\bibitem[Ritter et~al.(2018)Ritter, Botev, and Barber]{ritter2018scalable}
Hippolyt Ritter, Aleksandar Botev, and David Barber.
\newblock A scalable laplace approximation for neural networks.
\newblock In \emph{International Conference on Learning Representations}, 2018.

\bibitem[Schwöbel et~al.(2022)Schwöbel, Jørgensen, Ober, and van~der
  Wilk]{schwoebel2022layer}
Pola Schwöbel, Martin Jørgensen, Sebastian~W. Ober, and Mark van~der Wilk.
\newblock Last layer marginal likelihood for invariance learning.
\newblock In \emph{Proceedings of the Twenty Fifth International Conference on
  Artificial Intelligence and Statistics (AISTATS)}, 2022.

\bibitem[van Amersfoort et~al.(2021)van Amersfoort, Smith, Jesson, Key, and
  Gal]{amersfoort2021}
Joost van Amersfoort, Lewis Smith, Andrew Jesson, Oscar Key, and Yarin Gal.
\newblock Improving deterministic uncertainty estimation in deep learning for
  classification and regression.
\newblock \emph{CoRR}, abs/2102.11409, 2021.

\bibitem[van~der Ouderaa and van~der Wilk(2021)]{van2021learning}
Tycho~FA van~der Ouderaa and Mark van~der Wilk.
\newblock Learning invariant weights in neural networks.
\newblock In \emph{Workshop in Uncertainty \& Robustness in Deep Learning,
  ICML}, 2021.

\bibitem[van~der Pol et~al.(2020)van~der Pol, Worrall, van Hoof, Oliehoek, and
  Welling]{van2020mdp}
Elise van~der Pol, Daniel Worrall, Herke van Hoof, Frans Oliehoek, and Max
  Welling.
\newblock Mdp homomorphic networks: Group symmetries in reinforcement learning.
\newblock \emph{Advances in Neural Information Processing Systems}, 33, 2020.

\bibitem[van~der Wilk et~al.(2018)van~der Wilk, Bauer, John, and
  Hensman]{van2018learning}
Mark van~der Wilk, Matthias Bauer, ST~John, and James Hensman.
\newblock Learning invariances using the marginal likelihood.
\newblock In \emph{Advances in Neural Information Processing Systems}, pages
  9938--9948, 2018.

\bibitem[Wenzel et~al.(2020)Wenzel, Roth, Veeling, {\'S}wiatkowski, Tran,
  Mandt, Snoek, Salimans, Jenatton, and Nowozin]{wenzel2020good}
Florian Wenzel, Kevin Roth, Bastiaan~S Veeling, Jakub {\'S}wiatkowski, Linh
  Tran, Stephan Mandt, Jasper Snoek, Tim Salimans, Rodolphe Jenatton, and
  Sebastian Nowozin.
\newblock How good is the bayes posterior in deep neural networks really?
\newblock In \emph{International Conference on Machine Learning}, 2020.

\bibitem[Xiao et~al.(2017)Xiao, Rasul, and Vollgraf]{FashionMNIST}
Han Xiao, Kashif Rasul, and Roland Vollgraf.
\newblock Fashion-mnist: a novel image dataset for benchmarking machine
  learning algorithms, 2017.

\bibitem[Zagoruyko and Komodakis(2016)]{zagoryuko2016wide}
Sergey Zagoruyko and Nikos Komodakis.
\newblock Wide residual networks.
\newblock In \emph{{BMVC}}. {BMVA} Press, 2016.

\bibitem[Zhang et~al.(2018)Zhang, Sun, Duvenaud, and Grosse]{zhang2018noisy}
Guodong Zhang, Shengyang Sun, David Duvenaud, and Roger Grosse.
\newblock Noisy natural gradient as variational inference.
\newblock In \emph{International Conference on Machine Learning}, pages
  5852--5861, 2018.

\bibitem[Zhang et~al.(2019)Zhang, Dauphin, and Ma]{zhang2019fixup}
Hongyi Zhang, Yann~N Dauphin, and Tengyu Ma.
\newblock Fixup initialization: Residual learning without normalization.
\newblock In \emph{International Conference on Learning Representations}, 2019.

\bibitem[Zhou et~al.(2020)Zhou, Knowles, and Finn]{zhou2020meta}
Allan Zhou, Tom Knowles, and Chelsea Finn.
\newblock Meta-learning symmetries by reparameterization.
\newblock \emph{arXiv preprint arXiv:2007.02933}, 2020.

\end{thebibliography}
